\documentclass[review]{elsarticle}

\usepackage[english]{algorithm2e}
\usepackage{hyperref}
\usepackage{multirow}
\usepackage{color}
\usepackage{xcolor} 
\usepackage{caption}
\usepackage{subcaption}
\usepackage{balance}
\usepackage{url}
\usepackage{titletoc}
\usepackage{amsmath} 
\usepackage{amssymb}
\usepackage{commath}
\usepackage{lineno,hyperref}
\modulolinenumbers[1]
\usepackage{dirtytalk}
\usepackage{array}

\usepackage{url}
\usepackage{graphicx}
\usepackage{subcaption}
\usepackage{wrapfig}

\usepackage{multirow}
\usepackage{lscape}
\usepackage{enumitem}
\usepackage{booktabs}

\usepackage{amsmath,amsfonts,bm}

\usepackage{ulem}

\journal{Neural Networks (Accepted)}

\begin{document}

\begin{frontmatter}

\title{Distributed Associative Memory Network \\ with  Memory Refreshing Loss}


\author[mymainaddress1]{Taewon Park\corref{myequal}}
\ead{ptw7998@gmail.com}

\author[mymainaddress2]{Inchul Choi\corref{myequal}}
\ead{sharpic77@gmail.com}

\author[mymainaddress1,mymainaddress2]{Minho Lee\corref{mycorrespondingauthor}}
\ead{mholee@gmail.com}

\address[mymainaddress1]{Department of Artificial Intelligence, Kyungpook National University, 80, Daehak-ro, Buk-go, Daegu, Republic of Korea}
\address[mymainaddress2]{NEOALI, 80, Daehak-ro, Buk-go, Daegu, Republic of Korea}


\cortext[myequal]{Co-first authors with equal contribution}
\cortext[mycorrespondingauthor]{Corresponding author}

\begin{abstract}
Despite recent progress in memory augmented neural network (MANN) research, associative memory networks with a single external memory still show limited performance on complex relational reasoning tasks.
Especially the content-based addressable memory networks often fail to encode input data into rich enough representation for relational reasoning and this limits the relation modeling performance of MANN for long temporal sequence data.
To address these problems, here we introduce a novel Distributed Associative Memory architecture (DAM) with Memory Refreshing Loss (MRL) which enhances the relation reasoning performance of MANN. Inspired by how the human brain works, our framework encodes data with distributed representation across multiple memory blocks and repeatedly refreshes the contents for enhanced memorization similar to the rehearsal process of the brain.
For this procedure, we replace a single external memory with a set of multiple smaller associative memory blocks and update these sub-memory blocks simultaneously and independently for the distributed representation of input data. 
Moreover, we propose MRL which assists a task's target objective while learning relational information existing in data. MRL enables MANN to reinforce an association between input data and task objective by reproducing stochastically sampled input data from stored memory contents.
With this procedure, MANN further enriches the stored representations with relational information.
In experiments, we apply our approaches to Differential Neural Computer (DNC), which is one of the representative content-based addressing memory models and achieves the state-of-the-art performance on both memorization and relational reasoning tasks.
\end{abstract}

\begin{keyword}
Memory augmented neural network, relational reasoning, distributed representation, auxiliary loss, machine learning.
\end{keyword}

\end{frontmatter}


\section{Introduction}\label{sec:intro}
The essential part of human intelligence for understanding the story and predicting unobserved facts largely depends on the ability to memorize the past and reasoning for relational information based on the pieces of memory. 
In this context, research on artificial intelligence has focused on designing a human-like associative memory network that can easily store and recall both events and relational information from a part of the information.

In the classical research of associative memory, such as Hopfield network~\cite{hopfield1982neural}, the associative nature of biological memory is implemented by reconstructing patterns through the implicit iterative minimization of an energy function~\cite{hopfield1982neural, steinbuch1961lernmatrix, willshaw1969non, kanerva1988sparse}.
In this type of memory, there are two classes, auto-associative memory, and hetero-associative memory.
In the auto-associative memory, a stored pattern is retrieved based on a partially known or distorted version, and hetero-associative memory outputs a pattern that is different from the input pattern.
However, these energy-based memory models have fundamental limitations in adding capacity for more stored patterns and lack the means of modeling the higher-order dependencies that exist in real-world sequential data.
In deep neural network research, many approaches generally model high-order dependencies in input sequential data with memory systems, such as Long Short Term Memory~(LSTM)~\cite{hochreiter1997long} or memory augmented neural networks~(MANN).
Especially, the recent approach in MANN constructs an associative memory with a content-based addressing mechanism and stored both input data and its relational information to a single external memory.
MANN has already proven to be an essential component on many tasks which need long-term context understanding~\cite{weston2014memory, sukhbaatar2015end, graves2014neural, graves2016hybrid, gulcehre2018dynamic}.
Also, compared to recurrent neural networks, it can store more information from sequential input data and correctly recall desired information from memory with a given cue.
However, even with its promising performance on a wide range of tasks, MANN still has difficulties in solving complex relational reasoning problems~\cite{weston2015towards}.
Since the content-based addressing model implicitly encodes data items and their relational information into one vector representation, they often resulted in a lossy representation of relational information which is not rich enough for solving relational reasoning tasks.
To address such weakness, some researchers find relational information by leveraging interaction between memory entities with multi-head attention~\cite{palm2018recurrent, santoro2018relational}. Others focus on long sequence memorization performance of memory~\cite{trinh2018learning, le2018learning, munkhdalai2019metalearned}.
Another attempt to apply a self-attention to memory contents and explicitly encode relational information to a separate external memory~\cite{le2020self}.
However, all those models need to explicitly find relational information among memory entities with a high computational self-attention mechanism and have to repeatedly recompute it on every memory update.

In this research, we approach the same problem in a much simpler and efficient way which is inspired by how our brain and deep neural networks represent and restore information. We hypothesize that if we can encode input data into richer representations, any MANN can provide enhanced relation modeling performance without exhaustive self-attention-based relation searching. Based on this assumption, we find the weakness of the MANN model and facilitate it with the human brain-like mechanisms. One of the main weaknesses of conventional MANN is its lossy representation of relational information~\cite{le2020self}. In terms of content-based addressing memory, it can be caused by both a single memory-based representation and long-temporal data association performance.
Although MANN learns to correlate sequential events across time, its representation is not rich enough to reflect complex relational information that exists in input data. 
Therefore, for the enhanced relation learning, we focus on the richness of representation which implicitly embeds associations existing in input data.
For this purpose, we obtain insight from the fundamental representation principles of the human brain and deep neural network.
One of the reasons that deep learning has seen tremendous success is neural networks’ ability to learn rich distributed representations of data.
The concept of distributed representation is not only a fundamental principle of the neural network but also enables a neural network to learn diverse concepts and relational patterns that exist in input data~\cite{hinton1986distributed, zhang1994representation, plate1994distributed, paccanaro2001learning, bengio2013representation, takase2016composing}. The distributed representation is an well-known efficient method for learning both concepts and their binding to relational concepts of conceptual components. Furthermore, it is also known as how the biological brain stores and represents information. Motivated by these facts, we apply the distributed representation concept to the associative memory (i.e. content-addressable memory) blocks to provide a rich representation of relational information.
In this paper, we propose a novel associative memory architecture, Distributed Associative Memory (DAM), which is based on the mechanism of how the information is represented in the deep neural network~\cite{hinton1986distributed, zhang1994representation, plate1994distributed, paccanaro2001learning, bengio2013representation, takase2016composing} and our brains~\cite{lashley1950search, bruce2001fifty}.
In DAM, we replace the single external memory with multiple smaller sub-memory blocks and update those memory blocks simultaneously and independently.
The basic operations for each associative memory block are based on the content-based addressing mechanism of MANN, but its parallel memory architecture allows each sub-memory system to evolve over time independently.  Through this procedure, the input information is encoded and stored in distributed representations. The distributed representation is a concept that stems from how the brain stores information in its neural networks and well known for its efficiency and powerful representational diversity.
Furthermore, similar to the underlying insight of multi-head attention~\cite{vaswani2017attention}, our DAM model can jointly attend to information from different representation subspaces at different sub-memory blocks and is able to provide a more rich representation of the same common input data.
To retrieve rich information for relational reasoning, we apply a soft-attention-based interpolation to the diverse representations distributed across multiple memories.
Compared to other relation reasoning task-focused MANN models~\cite{santoro2018relational, le2020self}, DAM architecture does not adopt a self-attention mechanism to retrieve relational information from sequential input data. Therefore, it does not introduce quadratic computational complexity for pairwise comparison, and also does not require much modification when applied to any baseline MANN model. 
In spite of such simplicity, DAM effectively enhances the relation reasoning performance of MANN through the distributed representation from multiple associative memory blocks. 
Therefore, one of the main distinctions between our model and other associative memories lies in the fact that our model’s association comes from several distributed representations of input data rather than a single representation of the input.

Moreover, to enrich long-term relational information in the memory, we introduce a novel Memory Refreshing Loss (MRL) which fortifies the relational modeling ability of the memory and generally enhances the long-term memorization performance of MANN. The MRL forces the memory network to learn to reproduce the number of stochastically sampled input data only based on the stored memory contents.
As if, other associated pieces of memory are reminded together whenever a person recalls a certain event in his memory, the data reproducing task enables MANN to have better association and memorization ability for input data. In our brain mechanisms, a similar concept is maintenance rehearsal operation which is repeatedly verbalizing or thinking about a piece of information.
MRL is designed to reproduce a predefined percentage of input representations in the memory matrix on average and, while optimizing two different tasks at the same time, keep the balance between MRL and target objective loss by dynamically re-weighting each task~\cite{liu2006influence, cui2019class}.
Compared to other MANN researches, it is an novel approach to adopt auxiliary loss for long-term relational information refreshment. By combining unsupervised reconstruction loss with the representation enhancing memory architecture in a multi-task learning setting, we obtain additional performance advantages in the relational reasoning task.
 
By combining the above two approaches, DAM, and MRL, our architecture provides rich representation which can be successfully used for tasks requiring both memorization and relational reasoning.
We apply our architecture to Differential Neural Computer (DNC)~\cite{graves2016hybrid}, which is one of the representative content-based addressing memory, to construct novel distributed associative memory architecture with MRL. DNC has promising performance on diverse tasks but also known to be poor at complex relational reasoning tasks. In experiments, we show that our architecture greatly enhances both memorization and relation reasoning performance of DNC, and even achieves state-of-the-art records.
Furthermore, although our architecture does not rely on the self-attention mechanism to encode relational information in the input sequence, our model shows similar or better performance on relation reasoning tasks compared to other self-attention-based MANN models. Considering the most other state-of-the-art relation-seeking MANN models are adopting the self-attention mechanism with $O(n^{2})$ complexity, our architecture effectively provides comparable or superior relation reasoning performance only with $O(n)$ computational complexity.
The main contributions of the proposed architecture are summarized as follows.
\begin{itemize}

  \item
  We show the effectiveness of biological brain-inspired mechanisms (distributed representation and rehearsing procedure) on associative memory network performance. We show that the distributed representation efficiently contributes to modeling the relational information in input sequence and memory refreshing loss successfully fortifies it.
  
 \item
 Our architecture provides a novel way of retrieving relational information from sequential input data for MANN, without relying on a computationally expensive self-attention mechanism. Our framework is simple and easily applicable to any type of MANN model for further performance enhancement.

  \item
  We address the limitation of recurrent type MANN models on relation reasoning tasks. Our architecture applies the distributed representation concept to the collection of content-addressable memory blocks and successfully enriches the representation with relational information in the input sequence.

\end{itemize}

In the following, backgrounds on biological brain mechanisms and related works for the distributed memory and MANN are presented in Section \ref{sec:related},
and proposed architecture is illustrated in Section \ref{sec:proposed}. The
 experimental analysis and conclusion are shown in Section \ref{sec:experiments} and  \ref{sec:conclusion}.
 The possible challenges and future works are provided in Section \ref{sec:future.work}.

\section{Related Works} \label{sec:related}
\subsection{Biological Brain Mechanism}
Our memory architecture is mainly inspired by the information processing mechanisms of the human brain.
In our brain, forging new memories for facts and events, or retrieving information to serve the current task all depend on how information is represented and processed throughout the brain.
In many research~\cite{lashley1950search, bruce2001fifty, crick1983function, thompson1991memory}, there is already a broad consensus that distributed neural representations play a vital role in constructing and retrieving memories.
For example, the first indications of the distributed character of memory in the cerebral cortex are provided in Lashley’s~\cite{lashley1950search} neuropsychological experiments~\cite{fuster1998distributed}.
Also, in other researches of cognitive models, it is shown that the distributed associative memory is an efficient way for information representation and has the robustness to the noise, and helpful for enhancing reasoning performance of memory networks~\cite{kohonen1979storage, austin1987distributed, austin1996distributed, sommer2003models}.
This distributed representation concept has been widely applied for content-addressable memory, deep neural networks, automatic generalization, and adaptive rule selection~\cite{hinton1986distributed}.
Inspired by these researches, in our model, we apply the distributed representation concept to the multiple associative memory blocks to provide rich representation for enhanced reasoning.
Moreover, in psychology researches, it is well known that human memory can be enhanced by the repetitive rehearsal process of past information. Whether it is short-term or long-term memory, rehearsal can provide improved recall performance and working memory for the current task~\cite{rundus1980maintenance, greene1987effects}. Based on those researches, we design a new auxiliary loss function that is similar to the rehearsal process in our brain. Our loss function repetitively reconstructs some amount of previous input data based on the contents of memory while training for a target task. Since two different objectives are simultaneously trained with a multi-tasking learning setting,  this procedure is comparable to the psychological case study when a person is intentionally rehearsing some information while recognizing its use for a different task.
In this research, we experimentally show that such similarity effectively enhances the reasoning performance of memory network model, and also provides improved memorization performance.

\subsection{Neural Networks for Associative Memory}
There have been many research on associative memory network, from classical energy minimization based models to the memory augmented neural networks (MANN) approaches. These researches can be categorized as following with its relational modeling capability.

\paragraph*{Classical Associative Memory}
Conventional associative memory models are built upon the idea of reconstructing patterns with iterative energy minimization. This type of memory is described as an auto-associative memory when it reconstructs a previously stored pattern that mostly resembles the current pattern~\cite{hopfield1982neural, kohonen1972correlation}. Another variation is the hetero-associative memory (see e.g. \cite{kosko1988bidirectional}) where the retrieved pattern is different from the input pattern.
These memory models are simple and have similar connection architecture as the hippocampus of the brain. However, their capacity for stored patterns is limited and could not model the high-order dependencies in input data. The Boltzmann Machine~\cite{ackley1985learning} lifted some of these constraints (capacity limit) by introducing latent variables but at the cost of requiring slow reading and writing mechanisms.
There are also researches on the time-delayed input signal of neural networks in the real world applications. In such works, authors are mainly focused on addressing the neural network stability with respect to disturbance or uncertainty from real-world environment~\cite{chanthorn2020robust, rajchakit2020extended, rajchakit2021robust}. As an another type of approach, some works apply fractional-order neural networks and Bidirectional Associative Memory (BAM) for the time-delayed input of control systems~\cite{xu2019bifurcation, xu2021bifurcation, xu2021fractional}. However, these models are not capable of modeling the complex high order relational information that exists in sequential input data.
Compared to these works, our DAM architecture is initially designed to enhance relation modeling performance of any given MANN model with brain-inspired mechanisms. It has a simple distributed memory architecture that can be easily applied to any content-addressable MANN models to construct associative memory. Furthermore, our model adopts brain-inspired mechanisms, such as distributed representation and rehearsal process, to efficiently enrich representation of baseline MANN models and, finally, improves its relational information modeling performance.

\paragraph*{Multiple Memory based MANN}
In memory slot-based MANNs, the content-based addressing is implemented with a dynamic long-term memory which is composed of multiple memory slots~\cite{santoro2018relational, danihelka2016associative, henaff2017tracking, goyal2021recurrent}.
For multiple memory matrix-based models, researchers improve a single memory architecture by adding task-relevant information, asynchronous data input, and relational information to an additional memory matrix (e.g. dual memory)~\cite{le2020self, munkhdalai2017neural, le2018dual}.
Our DAM adopts multiple memory matrices for distributed representation. Compared to other approaches, distributed memory architecture is much simpler and shows better performance on the same problems.

\paragraph*{Memory Networks for Relational Reasoning} 
For relational reasoning, some MANN models explicitly found relational information by comparing their memory entities. Relational Memory Core~(RMC)~\cite{santoro2018relational} leverages interaction mechanisms among memory entities to update memory with relational information.
Self-attentive Associative Memory~(STM)~\cite{le2020self} adopts self-attention for memory contents and store relational information to separate relation memory. 
Compared to those methods, DAM provides relational information through diverse representations of input data and the long-term association performance of memory.

\paragraph*{Losses for Long-term Dependency}

For long-term memorization of the input pattern, \cite{munkhdalai2019metalearned} used a meta objective loss which forces a model to memorize input patterns in the meta-learning framework.
Also, for longer sequence modeling, \cite{trinh2018learning} adopted unsupervised auxiliary loss which reconstructs or predicts a sub-sequence of past input data.
Compared to \cite{trinh2018learning}, MRL does not rely on a random anchor point and the sub-sequence reconstruction rather enforces memorization of every past input data that are associated with a target task.
MRL focuses on enhancing data association while reproducing input representations, but also considering a balance with target objective loss by applying the dynamic weighting method for dual-task optimization.

\subsection{Differentiable Neural Computer}\label{sec:dnc}
We first briefly summarize DNC architecture which is a baseline model for our approaches.
DNC~\cite{graves2016hybrid} is a memory augmented neural network inspired by conventional computer architecture and mainly consists of two parts, a controller and an external memory.
When input data are provided to the controller, usually LSTM, it generates a collection of memory operators called as an interface vector~$\bm\xi_t$ for accessing an external memory. It consists of several \textit{keys} and \textit{values} for read/write operations and constructed with  the controller internal state~$\bm{h}_{t}$ as $\bm\xi_t = W_{\xi} \bm{h}_{t}$ at each time step $t$.
Based on these memory operators, every read/write operation on DNC is performed.

During the writing process, DNC finds a writing address,~$\bm{w}_t^w \in [0,1]^A$, where $A$ is a memory address size, along with writing memory operators, e.g. write-in \textit{key}, and built-in functions.
Then it updates write-in \textit{values},~$\bm{v}_t \in \mathbb{R}^L$, in the external memory,~$\bm{M}_{t-1} \in \mathbb{R}^{A \times L}$, along with erasing value,~$\bm{e}_t \in [0,1]^L$, where $L$ is a memory length size as follows:
\begin{equation}
\label{eq.original.write}
\bm{M}_t=\bm{M}_{t-1}\circ(\bm{E}-\bm{w}_t^{w}\bm{e}_t^\top)+\bm{w}_t^{w}\bm{v}_t^\top
\end{equation}
where $\circ$ denotes element-wise multiplication and $\bm{E}$ is $\bm{1}^{A \times L}$.

In the reading process, DNC searches a reading address,~$\bm{w}_t^{r,i} \in [0,1]^A$, for $R$ read heads, along with read memory operators, e.g. read-out \textit{key}.
Then, it reads out information from the external memory:
\begin{equation}
\label{eq.original.read}
\bm{r}_t^i = \bm{M}_t{\bm{w}_t^{r,i}}^\top
\end{equation}

Finally, the output is computed as $\bm{y}_t=W_y[\bm{h}_t;\bm{r}_t] \in \mathbb{R}^{d_o}$, where $\bm{r}_{t} = \{\bm{r}^i_{t} \in \mathbb{R}^L;1 \leq i \leq R\}$.
Through these operations, DNC can learn how to store input data and utilize stored information to solve a given task.
These whole mechanisms make DNC suitable for a general purposed memory augmented neural network.

\section{Proposed Method} \label{sec:proposed}
In this section, we introduce our two methods that improve both the memorization and relational reasoning ability of conventional DNC, a distributed associative memory architecture, and an MRL function. Then we analyze the computational overheads that come from the proposed method.
For a clear explanation, we illustrate DAM mechanism with a single read head case. For $R$ read head cases of DAM, the details are in \ref{appendix.implementation.detail}.

\begin{figure}[t]
\centering\includegraphics[width=\linewidth]{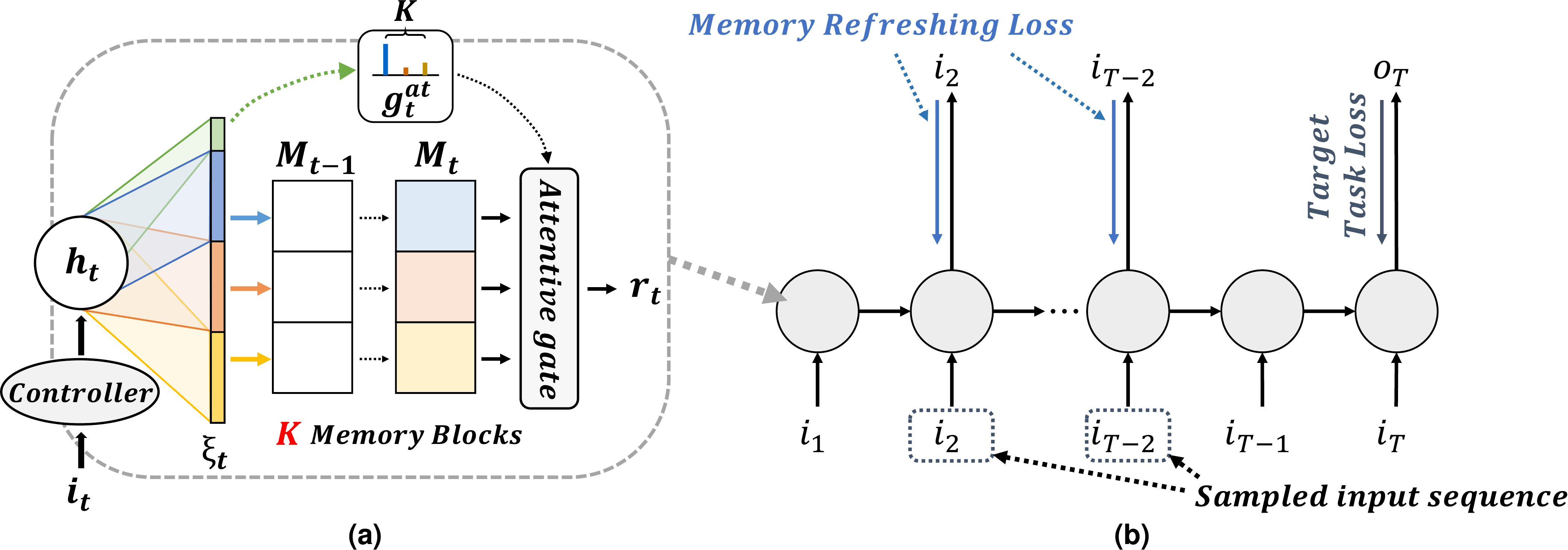}
\caption{(a) The DAM with $K$ sub-memory blocks (DAM-K) and attentive interpolation,~$g_t^{at}$. (b) Memory Refreshing Loss.
In the figure, $K$ denotes the number of memory blocks, and $h_t$, $\bm{M}_t$, $\bm\xi_t$, $g_{t}^{at}$, $\bm{i}_t$, and $\bm{o}_t$ represent controller internal state, external memory, interface vector, input, and output at time step $t$, respectively.}~\label{fig.proposed.method}
\end{figure}

\subsection{Distributed Associative Memory Architecture}
\label{mdma}

The distributed associative memory architecture consists of a controller network and $K$ associative memory blocks where each memory block is a content addressable memory similar to the original DNC~\cite{graves2016hybrid}.
Fig.~\ref{fig.proposed.method}(a) shows the overall read/write process of the proposed DAM.
For the writing operation, the controller of DAM produces multiple writing operator vectors for multiple memory blocks.
Each writing operator vector is used for the content-based addressing of one of the multiple memory blocks, and it is independent of other memory blocks.
Since it is produced based on the current input and previous hidden states of the controller,
it can independently store its own representation of the same input contents.
This writing process enables DAM to store the diverse representations of the same input data to multiple memory blocks with much flexibility.
Furthermore, for the reading process, all memory blocks are read at the same time and read values are interpolated with soft attention to produce single read-out information.
Through this attention-based reading process, DAM retrieves the most suitable information for the current task from representations distributed in the multiple memory blocks.
Based on these read\slash write operations, DAM learns how to store and retrieve the diverse representations of input data for different purposed tasks.
The following sections detail the main operations of our DNC based DAM model and its advantages in computational complexity.

\subsubsection{Controller for Multiple Associative Memory Blocks}
\label{controller}
At each time step $t$, the controller receives an external input,~$\bm{i}_{t}$, read-out of the previous time step,~$\bm{r}_{t-1}$, and previous hidden state of controller,~$\bm{h}_{t-1}$,
to update its current hidden state,~$\bm{h}_t$. After layer normalization, it produces an interface vector,~$\bm\xi_t \in \mathbb{R}^{K*(L*R+3L+3R+3)}$, which includes read and write parameters for multiple memory access.

\subsubsection{Write into Multiple Sub-Memory Blocks}
\label{write}

The multiple memory writing processes in our architecture are based on the content-based memory accessing mechanism of DNC. A single memory block is addressed and updated with the same procedure of DNC, and such single memory block updating is applied to all blocks independently at the same time.
As shown in Eq.~(\ref{eq.our.controller}), each memory block has its own interface vector relevant weight ~$W_{\xi,1},\cdots,W_{\xi,k}$,~where $k \in \{1,\cdots,K\}$. Theses weights are multiplied with a controller hidden state vector, $\bm{h}_{t}$, and used for memory operations of each independent memory block as following.
\begin{equation}
\openup 1ex
\label{eq.our.controller}
\bm\xi_t=[\bm\xi_{t,1},\cdots,\bm\xi_{t,K},\hat{g}_{t}^{at}] =[W_{\xi,1},\cdots,W_{\xi,K},W_{\xi,at}]\bm{h}_{t}
\end{equation}
where $\bm\xi_{t,k}$ is a interface vector for each memory block and $\hat{g}_{t}^{at}$ is an attentive gate at time~$t$.

Based on a writing operator obtained from $\bm\xi_{t,k}$, DAM updates input information into each memory block,~$\bm{M}_{t-1,k}$, independently and simultaneously, following Eq.~(\ref{eq.original.write}).
 That independent and simultaneous writing procedures of sub-memory blocks allow that our DAM learns to construct diverse representations for the same common input data.

The following attention-based reading process is designed to integrate representations distributed across sub-memory blocks, and it contributes to enrich representation for relational reasoning tasks.

\subsubsection{Read from Multiple Sub-Memory Blocks}
\label{read}

As in the writing process, DAM obtains a reading operator from $\bm\xi_{t,k}$, and computes reading address, ~$\bm{w}_{t,k}^{r} \in [0,1]^A$, for each memory block.
Based on those addresses, DAM reads values from each memory block and derives read-out value,~$\bm{r}_t \in \mathbb{R}^L$, from them, using a processed attentive gate,~$g_{t}^{at} \in [0,1]^K$, as follows:
\begin{equation}
\label{eq.read.out}
\bm{r}_{t} = \sum_{k=1}^{K} g_{t,k}^{at} \bm{M}_{t,k}^\top{\bm{w}_{t,k}^{r}}
\end{equation}
where $g_{t,k}^{at} = Softmax(\hat{g}_{t,k}^{at})$ for $k=1,\cdots,K$.

Compared to Eq.~(\ref{eq.original.read}) of DNC, this reading process integrates representations stored in multiple memory blocks with the attentive gate and enables DAM to learn to provides the most appropriate distributed representation for a target task.

\subsubsection{Computational Complexity of DAM Architecture}
\label{ccdam}

Our DAM is a general memory architecture that can be easily applied to any single external memory-based memory network. Since it simply constructs the distributed memory system based on the multiple sub-memory blocks, its memory read$/$write operations for each sub-memory block are almost the same as the baseline MANN model. Therefore, the overall computational complexity of DAM architecture is $\mathcal{O}(K\mathcal{M})$ and it increases with $K$, where $K$ is the number of memory blocks and $\mathcal{O}(\mathcal{M})$ is the computational complexity of the baseline memory operations.
Furthermore, in DAM, memory operations for K sub-memory blocks can be performed in parallel, which makes additional computational overhead negligible.
In experiments, our architecture shows enhanced relational reasoning performance compared to the baseline models even with a small $K$ value.

\subsection{Memory Refreshing Loss}
\label{memory.refreshing.loss}

To enhance the relation modeling performance of a memory network, we design a novel auxiliary task, Memory Refreshing Loss (MRL), which can further improve the memorization performance of any given MANN.
Our MRL function is inspired by the psychological case study on the rehearsal process of the human brain.
In the study, if a person repeatedly rehearses given words or numbers while knowing its use for the following task, the overall memory performance is enhanced~\cite{rundus1980maintenance, greene1987effects, souza2015refreshing, camos2017maintenance}.
Similarly, the main role of MRL task is forcing a memory network to reproduce sampled input data based on its memory content while training. When MRL task is trained with the main target task of the model in a multi-task learning setting, main task-related representation, and its encoded association can be further emphasized while training~\cite{caruana1997promoting, ben2003exploiting, alonso2017multitask, rei2017semi}.

First, we define a task-specific target objective function,~$\mathcal{L}^{task}$, of conventional MANN and MRL~$\mathcal{L}^{mr}_t$ as follows:
\begin{equation}
\label{eq.original.loss}
\mathcal{L}^{task} = \sum_{t=1}^{T} A(t) \ell_{task}(\bm{o}_t,\bm{y}_t)
\end{equation}
where $T$ is a whole sequence size, $A(t)$ is a function at time $t$, which indicates whether current phase is in  answer or not, if  its value is 1,  then $t$ is in answer phases (otherwise 0).  $\bm{o}_t$ is a target answer and $\ell_{task}(\cdot,\cdot)$ is a task target dependent loss function.

\begin{equation}
\label{eq.association.reinforcing.loss}
\mathcal{L}^{mr}_t = \ell_{mr}(\bm{i}_t,\bm{y}_t)
\end{equation}
where $\ell_{mr}(\cdot,\cdot)$ is an input sequence dependent loss function, and $\bm{i}_t$ is an input, $\bm{y}_t$ is an output at time step~$t$, respectively.

Our MRL function is defined to use a sampled input sequence as its target data as shown in Eq.~(\ref{eq.association.reinforcing.loss}), and this procedure leads the model to refresh given input information while it is learning the given task.
 The error measure for MRL is adopted based on the input item type or main task characteristic. In this research, we use cross-entropy loss or $L2$ loss depending on a given task.

As shown in Fig.~\ref{fig.proposed.method}(b), MRL forces a model to learn to reproduce sampled input sequences from stored representations in memory.
When sampling input data, each item of input sequence is sampled with Bernoulli trial with probability,~$p$, in which we call it as reproducing probability and it is defined as follows:
\begin{equation}
\label{eq.sample}
P(\alpha(t)=1) = 1 - P(\alpha(t)=0) = p
\end{equation}
where $\alpha(t)$ is an indicator function that represents sampling status at time $t$.

For an input sequence of length $n$, the series of Bernoulli trial-based samplings is the same as a Binomial sampling of the input sequence. Therefore, for any input sequence, on average, $np$ samples are reconstructed by MRL because an expected value of Binomial sampling is a product between trial probability, $p$, and the number of trials, $n$.
This random sampling policy prevents the model from learning to simply redirect given input to the output of the model at every time step.

When adding MRL to the task-specific target objective for multi-task learning, we also need a new strategy that can control the balance between MRL and original target task loss~\cite{jalali2019atrial}.
Since, as the number of the story input increases, the MRL can overwhelm the total loss of the model.
To prevent this loss imbalance problem, we apply a re-weighting method~\cite{cui2019class, liu2006influence}, which dynamically keeps the balance between the target task objective~$\mathcal{L}^{task}$ and MRL~$\mathcal{L}^{mr}$.
Moreover, we also introduce a scaling factor, $\gamma$, to ensure the main portion of training loss can be the original target objective function.
\begin{equation}
\label{eq.weight}
\gamma =
\left\{
	\begin{array}{ll}
		\hat{\gamma}  & \mbox{if }~\hat{\gamma} \geq 1, \\
		1 & \mbox{otherwise}.
	\end{array}
\right.
\end{equation}
where $\hat{\gamma} = \frac{\sum_{t=1}^{T}S(t)\alpha(t)}{\sum_{t=1}^{T}A(t)}$ and $S(t)$ is an indicator function which represents whether current time step $t$ is in the story phase or not.
Finally, the total loss for the training of proposed model follows:
\begin{equation}
\begin{gathered}
\label{eq.our.loss}
\mathcal{L} = \gamma\mathcal{L}^{task} + \sum_{t=1}^{T}\alpha(t)\mathcal{L}^{mr}_t
\end{gathered}
\end{equation}

From above two memory related tasks, $\mathcal{L}^{task}$ and $\mathcal{L}^{mr}$, while a model learns to reproduce input representations, target task-related representations and their association are further emphasized at the same time. As a result, MRL works as an auxiliary task that reinforces data association for a target objective.

\section{Experiments and Results} \label{sec:experiments}
We evaluate each of our main contributions, Distributed Associative Memory architecture~(DAM) and MRL, separately for ablation study, and show the performance of DAM-MR for complex relational reasoning tasks, such as bAbI, $N^{th}$ farthest task, and Convex hull task.
In all experiments, we adopt well-known neural network generalization techniques that are used in \cite{franke2018robust} for our baseline DNC model. The detailed parameter settings and adopted generalization techniques are shown in \ref{appendix.configuration.detail}.

\subsection{Distributed Associative Memory Architecture Evaluation}

The distributed memory architecture is evaluated in three aspects. First, we show the verification of the basic memory network capability of DAM  with Algorithmic tasks. Second, for the evaluation of memory efficiency in data association performance, DAM is configured to have a similar total memory size as a single memory model and evaluated with the Representation Recall task. Third, scalability experiments of DAM show the effect of the number of sub-memory blocks on the relation reasoning performance. In this experiment, we represent DAM architecture with $K$ sub memory blocks as DAM-$K$. The scalability experiments are performed with two settings, one is iteratively dividing a fixed total memory size to obtain multiple sub-memory blocks, the other is use a fixed sub-memory block size and adopting additional sub-memory blocks while increasing total memory size.

\begin{figure*}[t]
\centering
\begin{subfigure}[t]{.33\textwidth}
    \centering
    \includegraphics[width=\linewidth]{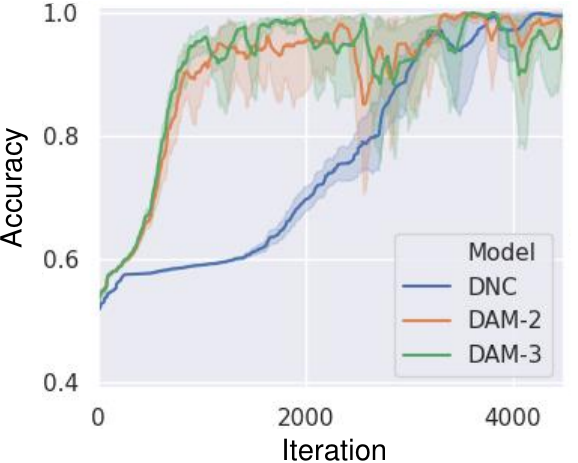}
    \caption{}
    \label{fig.algorithimic.copy}
  \end{subfigure}
  \hspace{14pt}
  \begin{subfigure}[t]{.33\textwidth}
    \centering
    \includegraphics[width=\linewidth]{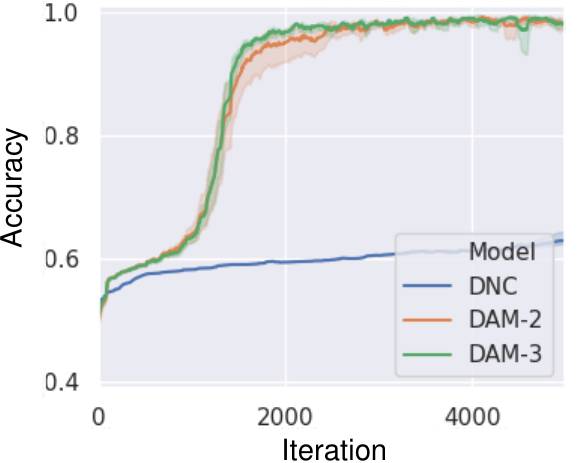}
    \caption{}
    \label{fig.algorithimic.ar}
  \end{subfigure}
\caption{Mean training curves on the algorithmic tasks which are (a) the copy task and (b) the associative recall task. The shadowed area shows a standard deviation of 10 trials.}
\label{fig.algorithmic}
\end{figure*}

\subsubsection{Algorithmic Tasks}
We show the effect of DAM on the basic memory network performance with the copy and the associative recall tasks from \cite{graves2014neural}.
The copy task is designed to show whether a model can store and recall arbitrary long sequential data correctly, and the associative recall task is intended to show whether a model can recall the information associated with a given cue by remembering temporal relation between input data. As shown in Fig.~\ref{fig.algorithmic}, simply adopting DAM architecture enhances the relation recall performance of the memory model. We can obtain more benefits by adding additional sub-memory blocks to DAM architecture (by increasing $K$, from 2 to 3), however, for the copy task, as shown in Fig.~\ref{fig.algorithmic}(a), the effect of the number of memory blocks is small because it is not a task designed for the evaluation of relation reasoning, rather focusing on simple memorization performance.

\begin{figure*}
  \begin{subfigure}[t]{.32\textwidth}
    \centering
    \includegraphics[width=\linewidth]{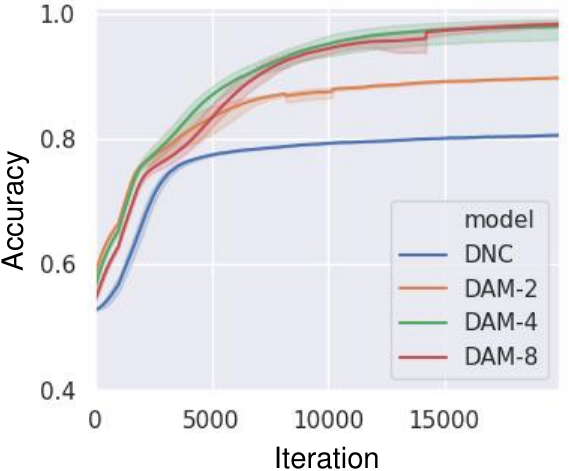}
    \caption{}
    \label{fig.rr.8segment}
  \end{subfigure}
  \hfill
  \begin{subfigure}[t]{.32\textwidth}
    \centering
    \includegraphics[width=\linewidth]{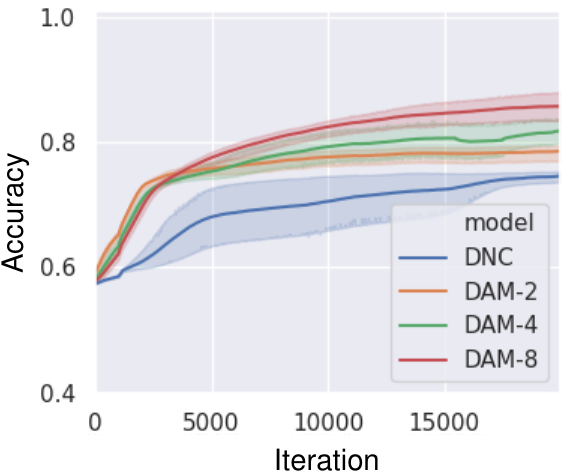}
    \caption{}
    \label{fig.rr.segment}
  \end{subfigure}
  \hfill
  \begin{subfigure}[t]{.33\textwidth}
    \centering
    \includegraphics[width=\linewidth]{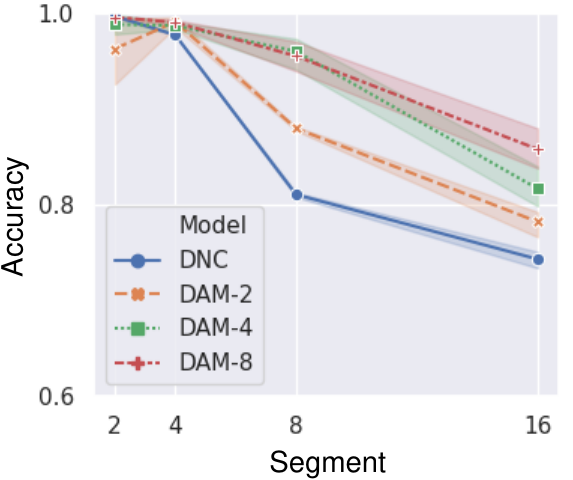}
    \caption{}
    \label{fig.rr.scalability}
  \end{subfigure}
\caption{Mean training curves for (a) 8 segment and (b) 16 segment on the Representation Recall task. (c) Mean accuracy of DAM models on the Representation Recall task. The shadowed area shows a standard deviation of 10 trials.}
\label{fig.rr}
\end{figure*}

\subsubsection{Representation Recall Task}

We design a new algorithmic task, called a Representation Recall (RR) task, which evaluates how much representation details a memory model can remember and recall from memory.
 This task uses randomly generated binary vectors as input sequences.
From the sequence, a binary vector is randomly selected and divided into $2N$ sub-parts. Among them, $N$ sub-parts are provided as a cue for a model to predict the rest of the sub-parts. In order to solve this task, the model is required to remember $\frac{2N!}{N!(2N-N)!}$ combinations of relations existing in each input, and therefore task complexity increases as $N$ increases.
To show the efficiency of a model with a fair comparison, we configure DAM by dividing the original single external memory into the group of $1/2$, $1/4$, and $1/8$ sized sub-memory blocks.
The mean training curves of DAM-$K$ ($K$=2, 4, and 8) are compared with the original DNC while increasing the task complexity $N$ as in shown Figs.~\ref{fig.rr}(a) and (b).
The result demonstrates that our proposed architecture learns the task much faster than other DNC based models, and also shows better accuracy and learning stability (smaller standard deviation in learning curve).
Furthermore, we compared the final accuracy of DAM-$K$($K$=2, 4, and 8) on RR task while increasing the task complexity from 2 to 16 segments.
As shown in Fig.~\ref{fig.rr}(c), if task complexity increases, the final accuracy inevitably degrades. However, DAM with more sub-memory blocks suffers less performance degradation. Although all of the models have  the same total memory size, a DAM model with more sub-memory blocks provides richer representation which includes more details of input.

\begin{figure*}[t]
  \begin{subfigure}[t]{.32\textwidth}
    \centering
    \includegraphics[width=\linewidth]{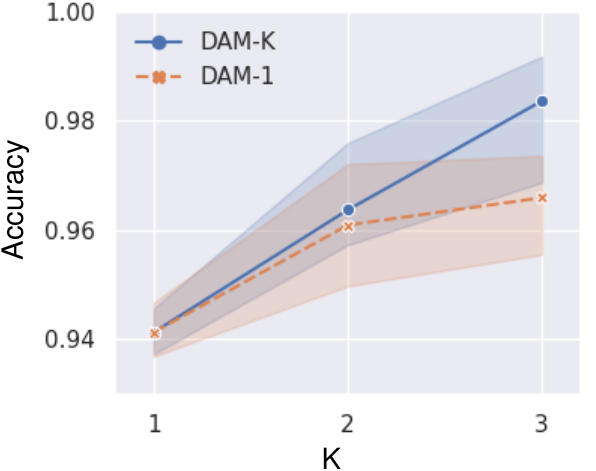}
    \caption{}
  \end{subfigure}
  \hfill
  \begin{subfigure}[t]{.31\textwidth}
    \centering
    \includegraphics[width=\linewidth]{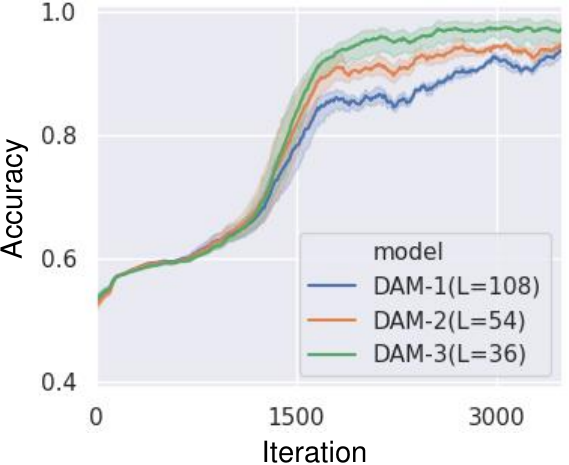}
    \caption{}
  \end{subfigure}
  \begin{subfigure}[t]{.32\textwidth}
    \centering
    \includegraphics[width=\linewidth]{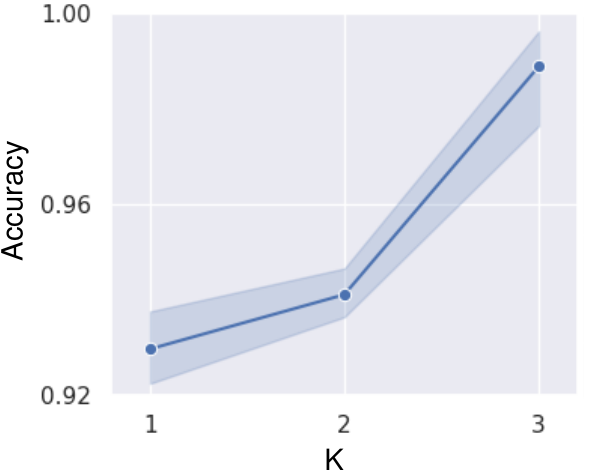}
    \caption{}
  \end{subfigure}
  \hfill
\caption{Scalability Experiments with Associative Recall task. (a) Scalability test while increasing single memory size, where $K$ is the number of memory blocks. (b) Training curve while iteratively dividing a fixed memory size, where $L$ is the memory length. (c) Scalability test while iteratively dividing a fixed memory size.}
\label{fig.ar.scalability}
\end{figure*}

\subsubsection{Scalability Experiments}
\paragraph*{Associative Recall task}
We perform a scalability experiment on the associative recall task with fixed memory address size, $A=16$. In order to evaluate the model efficiency of our DAM, we use two different configurations for the experiment. The first one is increasing $K$ while a single sub-memory block size is kept fixed. In this setting, DAM-$K$ is compared to the DAM-1 case model whose total memory size is set according to the corresponding DAM-$K$ model. In the second setting, we set every DAM-$K$ model to have the same fixed total memory size and iteratively divide the entire memory length, $L$, with $K$. Therefore, each of  DAM-1, 2, and 3 have its own $L$ size as $108$, $54$, and $36$, respectively. As shown in Fig.~\ref{fig.ar.scalability}(a), in the first case, DAM-$K$ provides more accuracy when total memory size increases. Fig.~\ref{fig.ar.scalability}(b) shows the training curves for the second setting, and in this figure, DAM architecture expedites the training speed of the model whenever sub-memory block size becomes smaller and the number of memory blocks increases. Fig.~\ref{fig.ar.scalability}(c) shows the final accuracy for the case of Fig.~\ref{fig.ar.scalability}(b). Similar to Fig.~\ref{fig.ar.scalability}(a), as we iteratively divide the memory size to obtain more sub-memory blocks, better accuracy is obtained if there is no information loss at a single sub-memory block. These experimental results from two different settings consistently show that a DAM applied model is always more efficient than a baseline MANN model when they have the same total memory size, and the performance of the DAM applied model is enhanced as the number of memory blocks is increased, if there is no information loss at the sub-memory block.

\begin{figure}[!t]
\centering\includegraphics[width=.34\linewidth]{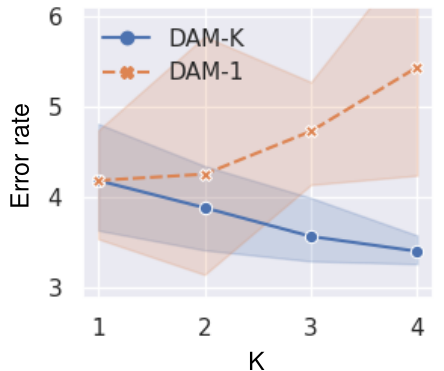}
\caption{Mean error rate of DAM models on the bAbI task, where $K$ is the number of memory blocks.}~\label{scalability}
\end{figure}

\begin{figure*}[t]
\centering

\begin{subfigure}[t]{.36\textwidth}
    \centering
    \includegraphics[width=\linewidth]{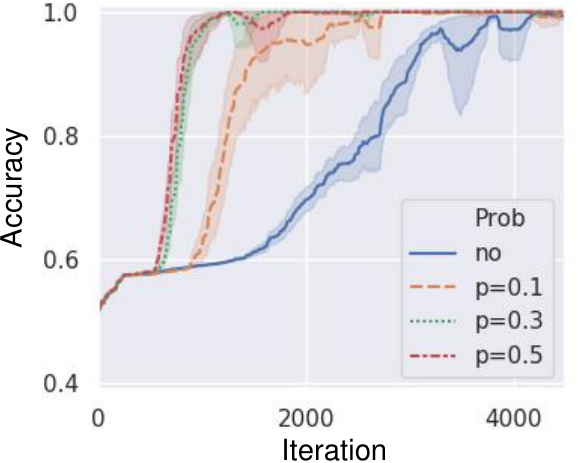}
    \caption{}
    \label{fig.mr.copy.dnc}
  \end{subfigure}
  \hspace{12pt}
  \begin{subfigure}[t]{.36\textwidth}
    \centering
    \includegraphics[width=\linewidth]{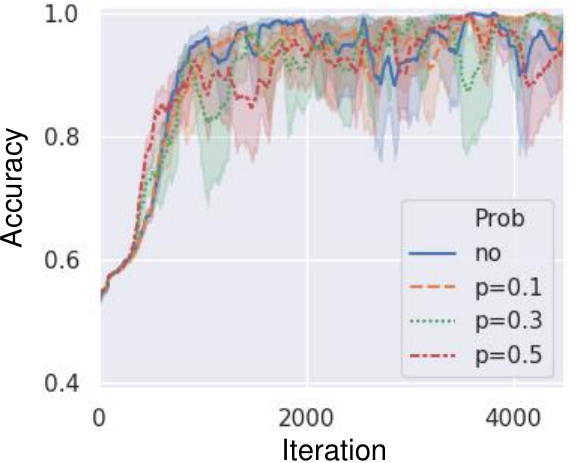}
    \caption{}
    \label{fig.mr.copy.dam3}
  \end{subfigure}
  
  \medskip
  
\begin{subfigure}[t]{.36\textwidth}
    \centering
    \includegraphics[width=\linewidth]{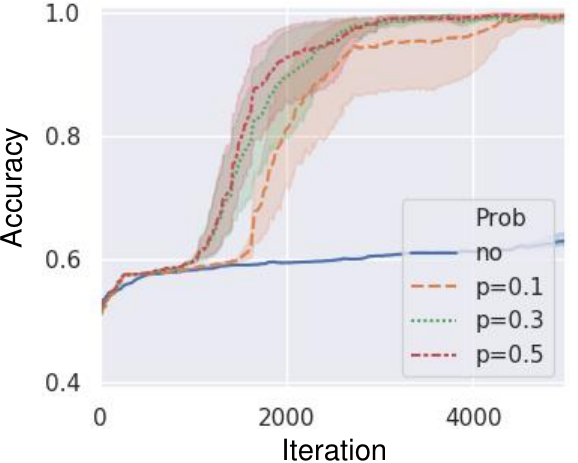}
    \caption{}
    \label{fig.mr.ar.dnc}
  \end{subfigure}
  \hspace{12pt}
  \begin{subfigure}[t]{.36\textwidth}
    \centering
    \includegraphics[width=\linewidth]{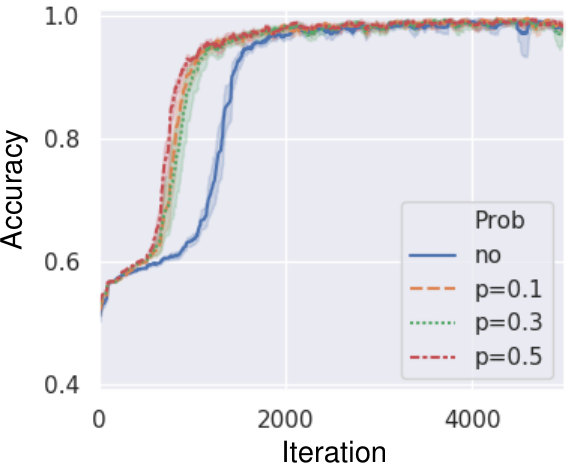}
    \caption{}
    \label{fig.mr.ar.dam3}
  \end{subfigure}
  
\caption{Mean training curves for different reproducing probability values at (a) DNC and (b) DAM-3 on the copy task. Mean training curves for different reproducing probability values at (c) DNC and (d) DAM-3 on the associative recall task. The shadowed area shows a standard deviation of 10 trials.}
\label{fig.mr.algorithmic}
\end{figure*}

\paragraph*{bAbI task}

For the evaluation of the scalability of distributed associative memory architecture without the effect of information loss at a sub-memory block,
we adopt a fixed size sub-memory block that has a larger length than half of the input size and then increase the number of sub-memory blocks to produce several models, DAM-2, 3, and 4.
We evaluate all model's performance with complex reasoning tasks, bAbI task, to show the effect of $K$ (representation diversity) on relational reasoning performance.
The bAbI task~\cite{weston2015towards} is a set of 20 different tasks for evaluating text understanding and reasoning, such as basic induction and deduction.
In Fig.~\ref{scalability}, the DAM-1 represents a baseline model that has a single external memory and includes modifications from~\cite{franke2018robust} for the generalization performance enhancement. For the comparison with DAM-$K$ ($K$=2, 3, and 4), we linearly increase its single external memory size. 
The overall graph shows that, as the degree of distribution increases, performance on bAbI tasks is also enhanced accordingly for both mean error rate and standard deviation of results. 
If we use more sub-memory blocks to further increase $K$, we can obtain gradual performance enhancement, which clearly shows the benefits of distributed associative memory architecture.

\subsection{Memory Refreshing Loss Evaluation}

To show the effect of MRL on MANN, we apply it to Algorithmic Tasks (Copy and Associative Recall task).
In Fig.~\ref{fig.mr.algorithmic}, we show the mean training curves according to the reproducing probability,~$p$, on the copy task, and the associative recall task, respectively. For DAM-MR, although we show only DAM3-MR, other configurations (DAM2-MR, DAM4-MR) have similar results.
As shown in Fig.~\ref{fig.mr.algorithmic}, the MRL function expedites the learning speed of models in most cases.
For the original DNC, it makes the training speed of the model much faster and it is further increased by the high reproducing probability on both tasks.
For DAM-MR, the MRL enhances the training speed of the models but  DAM is not sensitive to the change of reproducing probability. From those results, we can see that the effect of MRL is related to the property of a given task.

\subsection{DAM-MR Evaluation on Relational Reasoning Tasks}

As shown in the ablation study, each component of the proposed architecture has a significant impact on the original DNC performance. To show the performance of the whole combined model, DAM-MR, we compare our architecture to other DNC variations and attention-based MANN on following relational reasoning tasks. To support our argument on the relational information retrieving approach, we adopt recent memory network models which are applying extensive self-attention~\cite{le2020self} or multi-head attention for encoding relational information~\cite{santoro2018relational} as our counterparts.
Specifically, Relational Memory Core (RMC)~\cite{santoro2018relational} and Self-attentive Associative Memory (STM)~\cite{le2020self} apply the self-attention mechanism to its memory slots at each time step for relational information. When considering computational overhead required for relation retrieving, RMC and STM need quadratic computational complexity and it increases with the memory address size.
In contrast, our DAM-MR architecture needs almost the same computational complexity as the baseline MANN model (DNC), while providing superior relational information modeling performance.

\subsubsection{$N^{th}$ Farthest}
This task evaluates a model capacity for relational reasoning across time. It asks a model to find the
$N^{th}$ farthest vector from a given query vector, and this requires the memorization of relational information between vectors, such as distance, and sorting mechanism.
With this task, the long temporal relation modeling performance of a model can be demonstrated. 

\begin{table*}[t]
\caption{
Test accuracy [\%] on $N^{th}$ Farthest task.}
\begin{center}
\begin{tabular}{lccr}
\toprule
\large{\bf{Model}} & \large{\bf{Accuracy}} \\
\midrule
DNC~\cite{santoro2018relational}             & 25 \\
RMC~\cite{santoro2018relational}             & 91 \\
TPR~\cite{le2020self}             & 13 \\
STM~\cite{le2020self}             & 98 \\
\midrule
RMC-MR ($p=0.3$)    & 94 \\
DARMC4-MR ($p=0.3$)    & \textbf{98.2} \\
\midrule
DAM6-MR ($p=0.3$)   & 97.8 \\
\bottomrule
\label{table.nfar}
\end{tabular}
\end{center}
\end{table*}

Table~\ref{table.nfar} shows a comparison of the $N^{th}$ Farthest task results between our model and other MANN models which are designed for relational reasoning tasks. In the results, even though the original DNC can not solve the task at all, our DAM-MR shows surprisingly high performance on the task. Even compared to RMC~\cite{santoro2018relational}, which explicitly finds relational information from memory based on multi-head attention mechanism, DAM6-MR shows superior performance. For STM~\cite{le2020self}, our model shows a slightly lower accuracy. However, if we consider STM's self-attention computations for finding every possible relation with outer products, our DAM-MR is a quite simple and efficient architecture that does not introduce any explicit relation-seeking operations or high-order storage. Therefore, in the aspect of model efficiency, DAM-MR is showing a novel way of modeling relational information which is a promising alternative for the self-attention-based approach.
Moreover, if we apply our DAM architecture and MRL to the RMC, it even shows better performance than STM. Although RMC already has its own way of searching relational information from its memory, which overlaps with DAM's purpose, our DAM-MR provides further performance improvement on the task.
This result demonstrates the additional benefit of DAM architecture as a generally applicable design choice for MANN.

\begin{table*}[t]
\caption{Test accuracy [\%] on Convex hull task.}
\label{table.convexhull}
\begin{center}
\begin{small}
\begin{tabular}{lcc}
\toprule
\bf{Model}              & \bf{$N=5$} & \bf{$N=10$} \\
\midrule
LSTM~\cite{le2020self}  & 89.15 & 82.24 \\
ALSTM~\cite{le2020self} & 89.92 & 85.22 \\
DNC~\cite{le2020self}   & 89.42 & 79.47 \\
RMC~\cite{le2020self}   & 93.72 & 81.23 \\
STM~\cite{le2020self}   & 96.85 & 91.88 \\
\midrule
\textbf{DAM6-MR ($p=0.3$)}  & 95.6 & 89.8 \\
\textbf{DAM8-MR ($p=0.3$)}  & 95.4 & 90.5 \\
\midrule
\textbf{DARMC8-MR ($p=0.3$)} & \textbf{97.2} & \textbf{92.3} \\
\bottomrule
\end{tabular}
\end{small}
\end{center}
\end{table*}

\subsubsection{Convex hull task}
The convex hull task~\cite{vinyals2015pointer} is predicting a list of points that forms a convex hull sorted by coordinates.
The input list consists of $N$ points with 2D coordinates.
In this experiment, 
we train the model with $N \in [5,20]$ and test with $N =5,10$ cases.
The output is a sequence of 20-dimensional one-hot vectors representing
the features of the solution points in the convex-hull. As shown in Table~\ref{table.convexhull}, DAM-MR shows better accuracy than RMC~\cite{santoro2018relational} and similar performance with STM~\cite{le2020self}. Moreover, DARMC-MR, which is DAM applied RMC, shows even better performance than STM, which also demonstrates the effectiveness and generality of our DAM architecture.

\begin{table}[!htb]
    \begin{minipage}{.56\linewidth}
      \caption{The mean word error rate~[\%] for 10 runs of different DNC based models trained jointly on all 20 bAbI task.}
      \centering
        \begin{tabular}{lcc}
        \toprule
        \bf{Model} & \bf{Mean} & \bf{Best} \\
        \midrule
        DNC~\cite{graves2016hybrid}         & 16.7 \scriptsize{$\pm$ 7.6}   & 3.8 \\
        SDNC~\cite{rae2016scaling}          & 6.4 \scriptsize{$\pm$ 2.5}    & 2.9 \\
        rsDNC~\cite{franke2018robust}       & 6.3 \scriptsize{$\pm$ 2.7}    & 3.6 \\
        DNC-MD  & 9.5 \scriptsize{$\pm$ 1.6}    & n/a \\
        NUTM~\cite{Le2020Neural}            & 5.6 \scriptsize{$\pm$ 1.9}    & 3.3 \\
        \midrule
        \footnotesize{\textbf{DAM2-MR ($p=0.1$)}}              & \textbf{1.5} \scriptsize{$\pm$ 1.3}   & 0.16 \\
        \footnotesize{\textbf{DAM2-MR ($p=0.3$)}}              & 2.5 \scriptsize{$\pm$ 1.0}            & \textbf{0.14} \\
        \bottomrule
        \label{table.babi}
        \end{tabular}
    \end{minipage}%
    \quad
    \begin{minipage}{.4\linewidth}
      \centering
        \caption{The mean word error rate [\%] for best run of MAMN models trained jointly on all 20 bAbI tasks.}
        \begin{tabular}{lc}
        \toprule
        \bf{Model}                                  & \bf{Best} \\
        \midrule
        Transformer~\cite{dehghani2018universal}    & 22.1 \\
        UT~\cite{dehghani2018universal}             & 0.29 \\
        MNM-p~\cite{munkhdalai2019metalearned}      & 0.175 \\
        MEMO~\cite{Banino2020MEMO}                  & 0.21 \\
        STM~\cite{le2020self}                       & 0.15 \\
        \midrule
        \footnotesize{\textbf{DAM2-MR ($p=0.1$)}}                      & 0.16 \\
        \footnotesize{\textbf{DAM2-MR ($p=0.3$)}}                      & \textbf{0.14} \\
        \bottomrule
        \label{table.babi.best}
        \end{tabular}
    \end{minipage} 
\end{table}

\subsubsection{bAbI QA task}

The bAbI task~\cite{weston2015towards} is a set of 20 different tasks for evaluating text understanding and reasoning, such as basic induction and deduction.
Each task consists of stories for questions and correct answers for the questions, e.g.
$\mathit{Daniel~travelled~to~the~bathroom.}$
$\mathit{Mary}$
$\mathit{moved~to~the~office.}$
$\mathit{Where~is~Daniel?}$
$\mathit{bathroom}$.
In evaluation, a model is supposed to remember the story and recall related information to provide correct answer for the given questions.

Table~\ref{table.babi} shows experimental results on the bAbI task.
In this experimental result, our proposed model, DAM2-MR with $p=0.1$, shows the best mean performance on the bAbI task, among all other DNC based approaches.
These results demonstrate that our proposed architecture efficiently learns the bAbI task by using distributed associative memory architecture and memory refreshing loss.
Particularly, in Table~\ref{table.babi.best}, the best result of DAM2-MR records the state-of-the-art performance on the bAbI task, even compared to other types of recent MANN models.

\section{Conclusion} \label{sec:conclusion}
In this paper, we present a novel DAM architecture and an MRL function to enhance the data association performance of memory augmented neural networks.
The proposed distributed associative memory architecture stores input contents to the multiple sub-memory blocks with diverse representations and retrieves required information with soft-attention-based interpolation over multiple distributed memories.
We introduce a novel MRL to explicitly improve the long-term data association performance of MANN.
Our MRL is designed to reproduce the contents of associative memory with sampled input data and also provides a dynamic task balancing with respect to the target objective loss.
We implement our novel architecture with DNC and test its performance with challenging relational reasoning tasks.
The evaluation results demonstrate that our DAM-MR correctly stores input information and robustly recalls the stored information based on the purpose of the given tasks.
Also, it shows that our model not only improves the learning speed of DNC but reinforces the relation reasoning performance of the model.
Eventually, our DAM-MR significantly outperforms all other variations of DNC and shows the state-of-the-art performance on complex relation reasoning tasks, bAbI, even compared to other types of memory augmented network models.

\section{Possible Challenges and Future works} \label{sec:future.work}
In this paper, we experimentally show that DAM-MR is a simple and effective architecture for improving the relational reasoning performance of MANN, and it is easily applied to any MANN model. Because of such generality, DAM-MR can be used in any practical application domains where processing sequential information with high-order relations is important.
For example, Question and Answering tasks with multi-hop reasoning or long-term context sequences, Video action prediction, or reinforce learning-based applications. When applied to such practical application domains, one possible challenges could be the contextual length of sequential input data. As the contextual length for relational reasoning in input sequence becomes longer, it is more difficult to keep complex relations correctly in the memory and DAM-MR also needs trade off between additional computational overhead and the amount of performance enhancement it can obtain.

As future works, we are going to analyze the coordination mechanism between DAM and MR loss, and optimize our architecture so that it can be more adaptive to diverse types of other MANN models. 
In coordination mechanism, how MR Loss affects multiple sub-memory blocks while training, and how the task balancing strategy affects entire memory performance are challenging subject to be analyzed and we expect addressing these issues will contribute to designing a more powerful associative memory network for relational reasoning.

\section*{Acknowledgment}
This work was partly conducted by Center for Applied Research in Artificial Intelligence(CARAI) grant funded by Defense Acquisition Program Administration(DAPA) and Agency for Defense Development(ADD) (UD190031RD) (50\%). It was also supported by Electronics and Telecommunications Research Institute(ETRI) grant funded by the Korean government. [21ZS1100, Core Technology Research for Self-Improving Integrated Artificial Intelligence System](50\%).

\bibliography{mybibfile}

\newpage

\appendix
\section{Experiment Details}
\setcounter{figure}{0}
\setcounter{table}{0}

In experiments, for model optimization, we adopt an RMSprop optimizer with a momentum value as $0.9$ and epsilon as  $10^{-10}$.

\subsection{Model Configuration Details}
\label{appendix.configuration.detail}

\subsubsection{Distributed Associative Memory Architecture for DNC}

We configure hyper-parameters of our model with two types of settings, one for the comparison with other memory network models and the other for the scalability experiment of distributed memory (DM).
For the comparison with other models, we keep the almost same total memory size of DNC and divide it into smaller sub-memory blocks to construct a distributed memory system for a fair comparison. Also, we adjust the memory length size of DAM so that it has a similar amount of trainable parameters with other DNC variants.
Tables~\ref{table.copy.ar.babi} and~\ref{table.ditributed.nfar.convex} show our model hyper-parameters for each task, including a controller's internal state,~$d_h$, the number of read heads,~$R$, the number of memory blocks,~$K$, a memory address size,~$A$ and a memory length size,~$L$.

\begin{figure}[h]
\begin{center}
\centerline{\includegraphics[width=0.7\columnwidth]{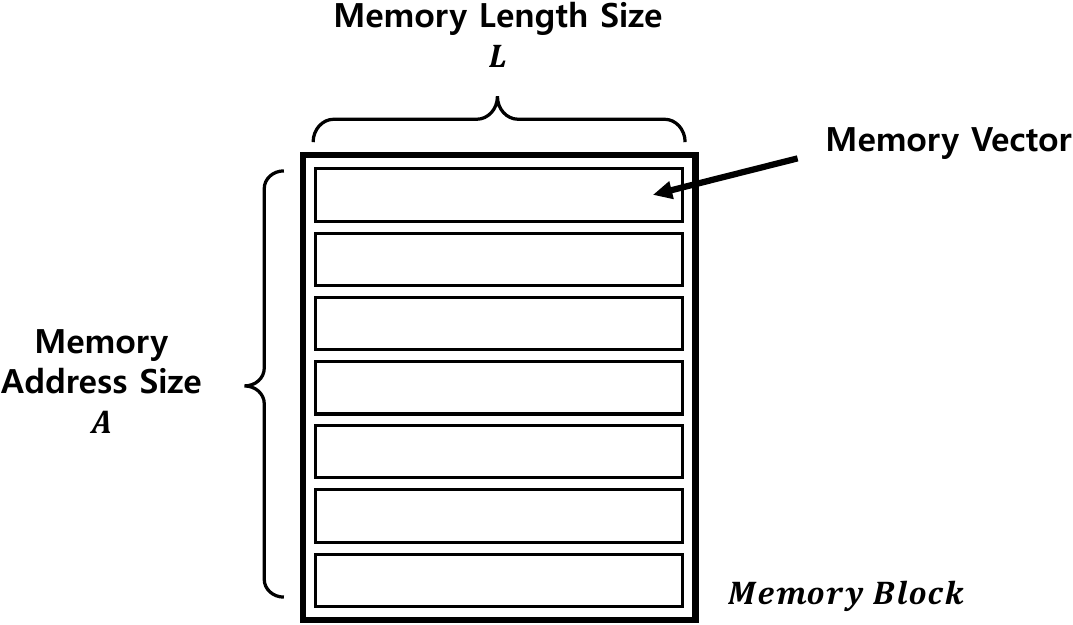}}
\caption{Composition of a memory block.}
\label{fig.block.}
\end{center}
\end{figure}

\begin{table}[!htb]
    \begin{minipage}{\linewidth}
      \caption{Model hyper-parameters for the copy task, associative recall task, and bAbI task.}
        \label{table.copy.ar.babi}
        \begin{center}
        \begin{footnotesize}
        \begin{tabular}{cccccccc}
        \toprule
        \multirow{2}{*}{Hyper parameter} & \multicolumn{2}{c}{Copy} & \multicolumn{2}{c}{Associative Recall} & \multicolumn{2}{c}{bAbI QA} \\
        \cmidrule(lr){2-3} \cmidrule(lr){4-5} \cmidrule(lr){6-6}
        & DNC & DAM-K & DNC & DAM-K & DAM-2 \\
        \midrule
        $d_h$ & 128 & 128 & 128 & 128 & 256  \\
        $R$ & 1 & 1 & 1 & 1 & 4 \\
        $K$ & 1 & \{2, 3\} & 1 & \{2, 3\} & 2\\
        $A$ & 64 & 64 & 32 & 32 & 128 \\
        $L$ & 36 & 36 & 36 & 36 & 48 \\
        \midrule
        Memory Capacity & $2.3$\textmd{K} & \{$4.6$\textmd{K}, $6.9$\textmd{K}\} & $1.1$\textmd{K} & \{$2.3$\textmd{K}, $3.4$\textmd{K}\} & $12.3$\textmd{K} \\
        Total Parameters & $0.11$\textmd{M} & \{$0.13$\textmd{M}, $0.15$\textmd{M}\} & $0.11$\textmd{M} & \{$0.13$\textmd{M},  $0.15$\textmd{M}\} & $0.79$\textmd{M} \\
        \bottomrule
        \end{tabular}
        \end{footnotesize}
        \end{center}
    \end{minipage}%
    \vspace{1cm}
    \begin{minipage}{\linewidth}
      \caption{Model hyper-parameters for the representation recall task, $N^{th}$ farthest task, and the convex hull task.}
        \label{table.ditributed.nfar.convex}
        \begin{center}
        \begin{footnotesize}
        \begin{tabular}{cccccc}
        \toprule
        \multirow{2}{*}{Hyper parameter} & \multicolumn{2}{c}{Representation Recall} & \multicolumn{1}{c}{$N^{th}$ Farthest} & \multicolumn{1}{c}{Convex Hull} \\
        \cmidrule(lr){2-3} \cmidrule(lr){4-4} \cmidrule(lr){5-5}
         & DNC & DAM-K & DAM-6 & DAM-K \\
        \midrule
        $d_h$ & 128 & 128 & 1024 & 256 \\
        $R$ & 1 & 1 & 4 & 4 \\
        $K$ & 1 & \{2, 4, 8\} & 6 & \{6, 8\} \\
        $A$ & 32 & 32 & 16 & 20 \\
        $L$ & 256 & $256/K$ & 128 & 64 \\
        \midrule
        Memory Capacity & $8.2$\textmd{K} & $8.2$\textmd{K} & $12.3$\textmd{K} & \{$7.7$\textmd{K}, $10.2$\textmd{K}\} \\
        Total Parameters & $0.38$\textmd{M} & \{$0.31$\textmd{M}, $0.27$\textmd{M},  $0.26$\textmd{M}\} & $12.6$\textmd{M} & \{$1.5$\textmd{M}, $1.7$\textmd{M}\} \\
        \bottomrule
        \end{tabular}
        \end{footnotesize}
        \end{center}
    \end{minipage} 
\end{table}

\begin{table*}[t]
\caption{Model hyper-parameters for scalability evaluation on the bAbI task.}
\label{table.scalability}
\begin{center}
\begin{small}
\begin{tabular}{ccccccc}
\toprule
Hyper parameter & DAM-1 & DAM-2 & DAM-3 & DAM-4\\
\midrule
$d_h$ & 256 & 256 & 256 & 256 \\
$R$ & 4 & 4 & 4 & 4 \\
$K$ &1 & 2 & 3 & 4 \\
$A$ & 192 & 128 & 128 & 128 \\
$L$ & 64 & 48 & 48 & 48 \\
\midrule
Memory Capacity & $12.3$\textmd{K} & $12.3$\textmd{K} & $18.4$\textmd{K} & $24.6$\textmd{K} \\
Total Parameters & $0.80$\textmd{M} & $0.79$\textmd{M} & $0.88$\textmd{M} & $0.97$\textmd{M} \\
\bottomrule
\end{tabular}
\end{small}
\end{center}
\end{table*}

For the scalability evaluation of DAM, we set the memory length of
a single sub-memory block,~$L$, same as the memory block size of DAM-2.
The size smaller than this causes the information loss because of too small memory matrix size compared to the input length. In this configuration, we increase the degree of distribution by increasing the number of sub-memory blocks.
Since we fix the single memory block size, the total memory size increases linearly with $K$.
Table~\ref{table.scalability} shows the details of configuration.

\subsubsection{Distributed Associative Memory Architecture for RMC}

\begin{figure}[h]
\centering\includegraphics[width=\linewidth]{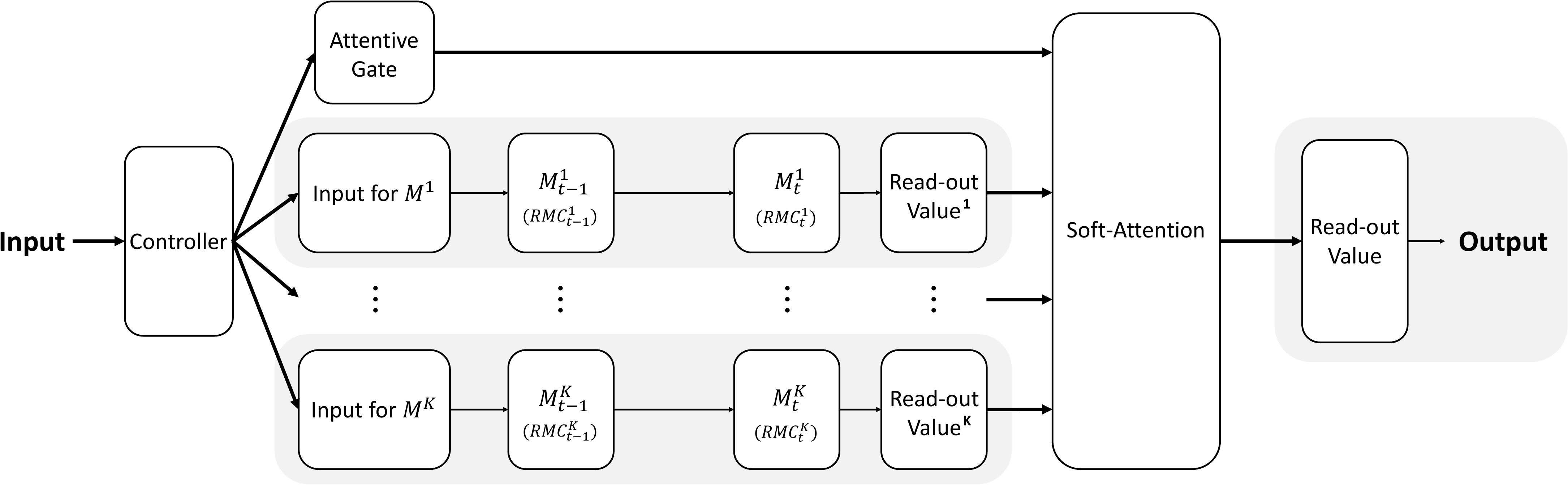}
\caption{The Distributed Associative Memory architecture applied RMC (DAMRMC). Each sub memory block can be replaced by any MANN model.}
\label{fig.generalized.DAM.architecture}
\end{figure}

To show the effectiveness and generality of our DAM architecture, we naively apply DAM architecture to RMC~\cite{santoro2018relational} by constructing a memory system with the collection of several RMC blocks, as shown in Fig.~\ref{fig.generalized.DAM.architecture}.
By treating a single RMC memory network as a single sub memory block, we compose a memory architecture with $K$ memory blocks.
For DAM operations, we add a controller network (feed-forward network) to provide input value to the multiple memory blocks, and for the memory read-out, integrate the read-out value of each memory block with attentive-gate based soft-attention mechanism.
We set the hyper-parameters of each RMC memory block as 8 memory slots with 1,536 total units and 6 heads for the $N^{th}$ Farthest task, and 8 memory slots with 2,048 total units and 8 heads for the convexhull task.

\subsection{Experimental Task Explanation}

In this subsection, we introduce our experimental setup and how to conduct experiments on each task.

\subsubsection{Algorithmic Task Description}

The experiments on algorithmic tasks are repeated 10 times with a batch size of $16$ and a learning rate of $10^{-4}$, and training iterations of $20$\textmd{K} on Representation Recall task and training iterations of $10$\textmd{K} on Copy and Associative Recall tasks.
We evaluate the performance on the algorithmic tasks based on the accuracy which is defined as $L1$ norm between model outputs and targets.
In each task, an input flag is provided with story~(input) data which are required to produce an answer.
After story input, query data along with the output flag are provided to the model.
Based on the input and stored information, the model is required to predict an correct answer.
For training, we randomly construct each task's data at each iteration as the following configurations.

\paragraph*{Representation Recall Task.}
In this task, $L_i$ binary vectors, each has $W$ length, are randomly generated and are provided to the model along with an input flag.
Here, we divide each input binary vector into $2N$ segments and use the half of them,~$N$, as a cue vector.
For an answer phase, $L_c$ cue vectors are constructed by random sampling from $L_i$ binary vectors with replacement and $N$ segments without replacement.
The network has to predict the remaining $N$ segments when each cue vector is provided.
We use $L_i=8$, $W=64$, $N \in \{2, 4, 8\}$, and $L_c \in [8, 16]$.
Therefore,  $d_i$ and $d_o$ are $64$ and $32$, respectively.

\paragraph*{Copy Task and Associative Recall Task.}
\label{detail.copy.ar}
In the copy task~\cite{graves2014neural}, $L_i$ binary vectors, each has $W$ length, are randomly generated and are provided to the model along with an input flag.
After receiving an output flag, the model is required to sequentially produce the same $L_i$ binary vectors as an output.
We use $W=8$ and the number of binary vectors, $L_i \in [8, 32]$, is randomly chosen at each iteration.
In the associative task~\cite{graves2014neural}, we define an item as a sequence of binary vectors with width $W$, and during the input phase, $L_i$ input items, each consists of $N_i$ binary vectors, are provided to a model along with an input flag.
Subsequently, a query item, which is randomly chosen from an input item sequence, is provided to the model along with an output flag.
In this phase, the model is required to predict a subsequent item that has placed right after the query item in the input item sequence.
We use $W=8$, $N_i=3$, and the number of items, $L_i \in [2, 8]$, is randomly chosen at each iteration.
Therefore, $d_i$ and $d_o$ are $10$ on both tasks.

\subsubsection{bAbI Task}

Our experiments on bAbI task are repeated 10 times with a batch size of $32$ and a learning rate of $[1,\underbar{3},10] \times 10^{-5}$, and we fine-tuned a model with the learning rate of $10^{-5}$ and training iterations of $10$K.
We compose experimental set on the bAbI task with 62,493 training samples and 6,267 testing samples and adopt fixed training iterations of about $0.1$\textmd{M}.
We use a word embedding, which is a general method in a natural language processing domain, with an embedding size of $64$.
Thus, $d_i$ is $64$ and $d_o$ is $160$.
We evaluate the performance on the bAbI task based on the word error rate for answer words in the test samples.

\paragraph*{Pre-processing on bAbI Task Dataset}

The bAbI task dataset~\cite{weston2015towards} consists of many different types of sets that share similar properties.
Among them, we only use en-10\textmd{K} set to fairly compare the performance with other researches.
In data pre-processing, any input sequences which have more than 800 words are all excluded from the training dataset as in \cite{franke2018robust} for computational efficiency.
After the exclusion, we construct our dataset with 62,493 training samples and 6,267 testing samples.
We remove all numbers from input data and split every sentence into words, and convert every word to its corresponding lower case.
After these processes, the whole vocabulary composed of 156 unique words and four symbols, which are '[PAD]', `?', `.', and `-'.
The `-' symbol represents that the model has to predict an answer word at the current time step.
Therefore, answer words are replaced to '-' symbol from input data.
The bAbI task dataset is available on \url{http://www.thespermwhale.com/jaseweston/babi/tasks_1-20_v1-2.tar.gz}.

\subsubsection{$N^{th}$ Farthest task}
We use the same task configuration used in \cite{santoro2018relational}, which uses Adam Optimizer with a batch size of 1,600 and an initial learning rate of $1e^{-4}$, consists of  eight 16-dimensional input vectors, and add 4-layers of MLP with ReLU activation function to the output layer.
We set DAM configurations as shown in Table~\ref{table.ditributed.nfar.convex}.

\subsubsection{Convex hull task}
We use the same task configuration used in \cite{le2020self}, which adopts RMSProp Optimizer with a batch size of 128 and an initial learning rate of $1e^{-4}$. 
We add 2-layers of MLP with ReLU activation function (each layer has 256 units) to the output layer and set DAM configurations as shown in Table~\ref{table.ditributed.nfar.convex}.

\section{Additional Analysis of Memory Refreshing Loss}

\setcounter{table}{0}

Proposed Memory Refreshing Loss~(MRL) chooses a subset of input sequence based on a stochastic sampling with trial probability,~$p$, which is called a reproducing probability.
It adaptively decides the number of sampled input story,~$n^\prime$, according to the story sequence length,~$n$, by adjusting this reproducing probability. Since it is a binomial sampling with independent trials, the expected sampled story length is as shown in Eq.~(\ref{eq.ARL.expectation}).
Therefore, it can consistently enhance the memorization performance of memory with the reproducing probability of $p$.
\begin{equation}
\label{eq.ARL.expectation}
E[n^\prime] = n p
\end{equation}

We use an input sequence dependent loss function,~$L^{mr}$, of each task as shown in Table~\ref{table.loss.list}.
We adopt $Cross Entropy$ loss function for the Algorithmic, Representation Recall, and bAbI tasks since their input sequences are the binary vectors or one-hot encoding vectors, whereas $L2$ loss function is adopted to $N^{th}$ farthest task and the convex hull task because their input data are continuous values.

\begin{table*}[h]
\caption{The input data dependent loss function,~$L^{mr}$, for each task.}
\label{table.loss.list}
\begin{center}
\begin{footnotesize}
\begin{tabular}{l|ccccc}
\toprule
\normalsize{\bf{Task}}     &   Algorithmic      & Representation Recall     & bAbI      & $N^{th}$ Farthest     & Convex hull \\
 \cmidrule(lr){1-1} \cmidrule(lr){2-4} \cmidrule(lr){5-6}
\normalsize{\bf{Loss}}     &  \multicolumn{3}{c}{$Cross Entropy$}  &   \multicolumn{2}{c}{$L2$} \\
\bottomrule
\end{tabular}
\end{footnotesize}
\end{center}
\end{table*}

\section{Visualization of Distributed Memory Operation (Attentive Gate)}

\setcounter{figure}{0}

\begin{figure*}[h]
\centering

\begin{subfigure}[t]{.8\textwidth}
    \centering
    \includegraphics[width=\linewidth]{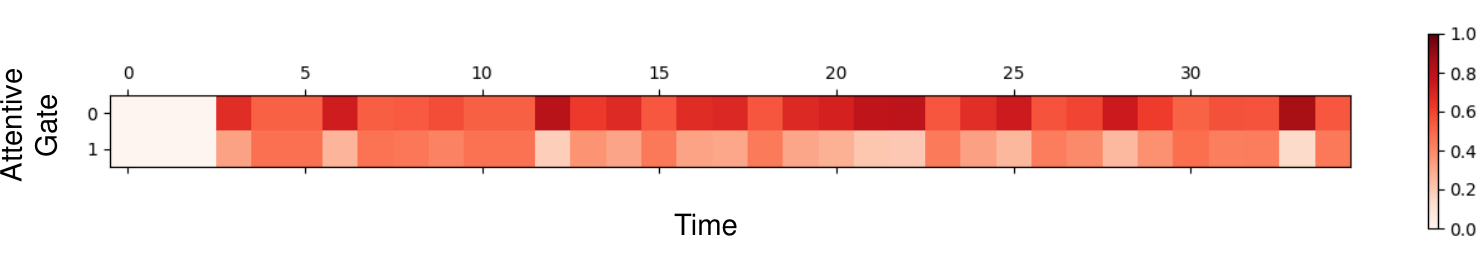}
    \caption{}
    \label{fig.attentive.gate.dam2.8}
  \end{subfigure}
  
  \medskip
  
\begin{subfigure}[t]{.8\textwidth}
    \centering
    \includegraphics[width=\linewidth]{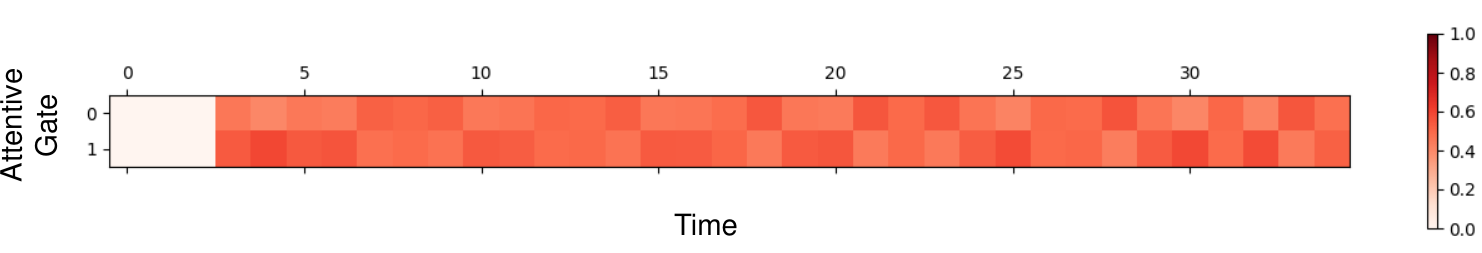}
    \caption{}
    \label{fig.attentive.gate.dam2.16}
  \end{subfigure}
  
\caption{The activated attentive gate of DAM-2 according to time step on Representation Recall task. (a) $8$ sub-parts. (b) $16$ sub-parts. In figure, the value indicates how much include each memory content when the model generates output at each time step.}
\label{fig.attentive.gate.dam2}
\end{figure*}

\begin{figure*}[h]
\centering

\begin{subfigure}[t]{.8\textwidth}
    \centering
    \includegraphics[width=\linewidth]{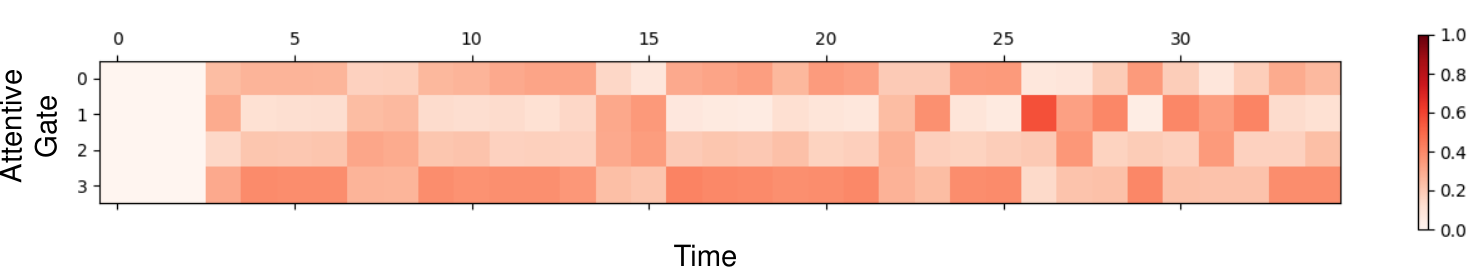}
    \caption{}
    \label{fig.attentive.gate.dam4.8}
  \end{subfigure}
  
  \medskip
  
\begin{subfigure}[t]{.8\textwidth}
    \centering
    \includegraphics[width=\linewidth]{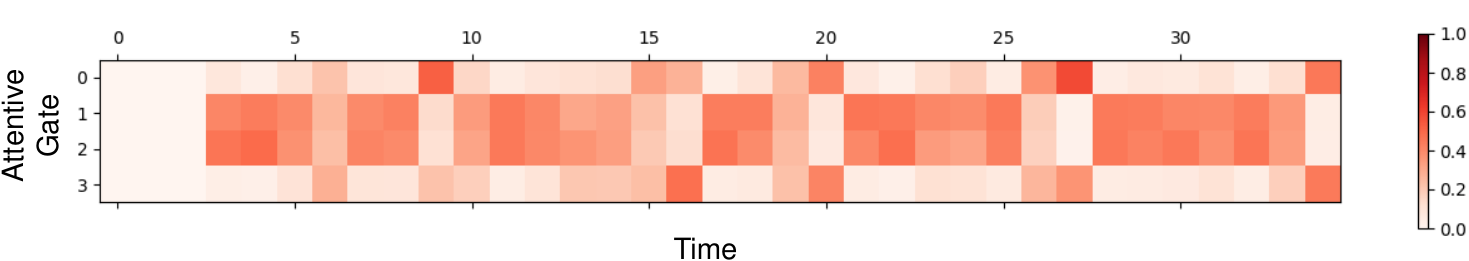}
    \caption{}
    \label{fig.attentive.gate.dam4.16}
  \end{subfigure}
  
\caption{The activated attentive gate of DAM-4 according to time step on Representation Recall task. (a) $8$ sub-parts. (b) $16$ sub-parts. In figure, the value indicates how much include each memory content when the model generates output at each time step.}
\label{fig.attentive.gate.dam4}
\end{figure*}

Fig.~\ref{fig.attentive.gate.dam2} and Fig.~\ref{fig.attentive.gate.dam4} show relative weights of multiple memory blocks (DAM-2 and 4) obtained from attentive gate  while performing Representation Recall~(RR) task.
During RR task, the model has to predict missing sub-parts when given other parts as a clue. The figures show how much each sub memory block is referenced when the model is predicting sub-parts.
As shown in the above figures, for given tasks, the model effectively integrates well-distributed information from multiple memory blocks and utilizing all memory blocks for information retrieval.

\section{Additional Results on bAbI task}

\setcounter{figure}{0}

\begin{figure}[h]
\begin{center}
\centerline{\includegraphics[width=0.5\columnwidth]{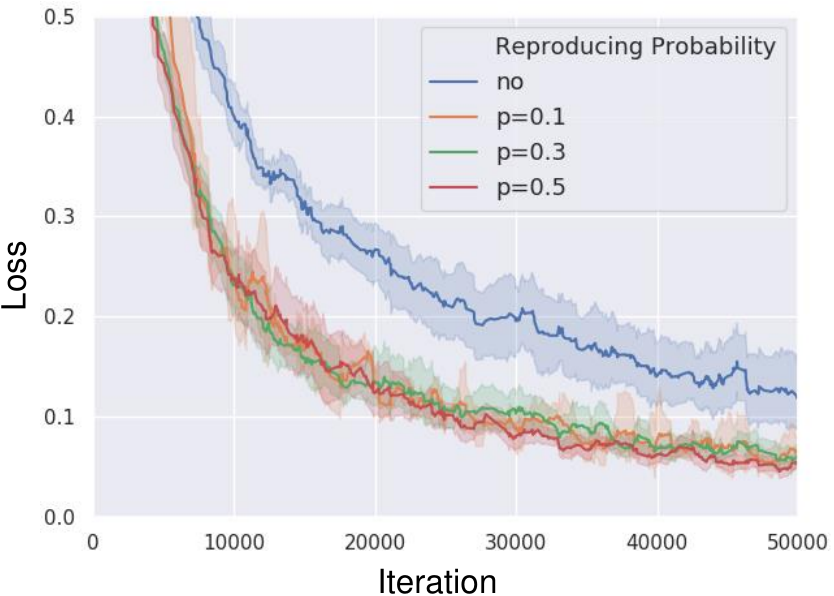}}
\caption{Mean convergence curves of DAM2-MR for different reproducing probability on the bAbI task. The shadowed area shows a standard deviation of 10 trials.}
\label{fig.babi.loss.}
\end{center}
\end{figure}

\section{Additional Experimental Analysis on DAM-MR}

In the experimental section of our paper, we set up each block size as \{$A=64, L=36$\} for the copy task and \{$A=32, L=36$\} for the associative recall task. To show the model performance with a smaller number of memory slots and memory block length, we show the additional experimental analysis.

\subsection{Experiments with Smaller Memory block Address Size}

\setcounter{figure}{0}

\begin{figure*}[t]
\centering
\begin{subfigure}[t]{.4\textwidth}
    \centering
    \includegraphics[width=\linewidth]{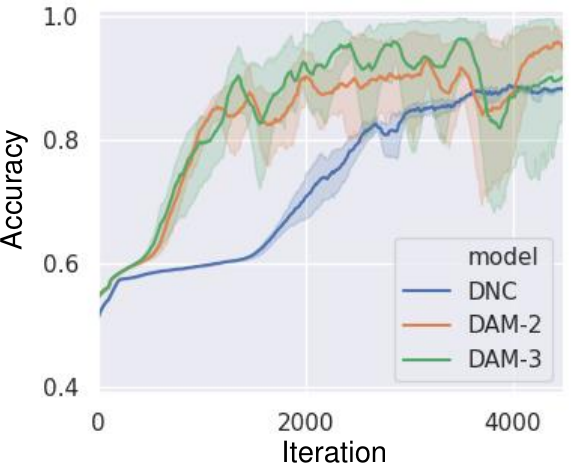}
    \caption{}
    \label{fig.reduced.algorithimic.copy}
  \end{subfigure}
  \hspace{12pt}
  \begin{subfigure}[t]{.4\textwidth}
    \centering
    \includegraphics[width=\linewidth]{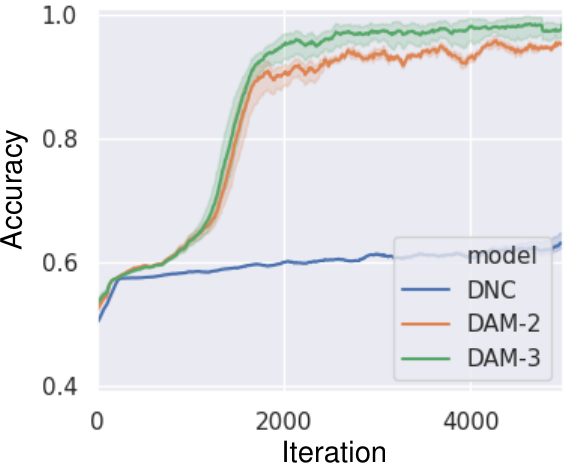}
    \caption{}
    \label{fig.reduced.algorithimic.ar}
  \end{subfigure}
\caption{Mean training curves on the algorithmic tasks with $A=16$, which are (a) the copy task and (b) the associative recall task. The shadowed area shows a standard deviation of 5 trials.}
\label{fig.reduced.algorithmic}
\end{figure*}

In our previous Algorithmic experiments, for the copy task, the number of input binary vectors are randomly chosen from $L_i \in [8, 32]$, and the memory address size is set up as $A=64$. For the associative recall task, the number of items is chosen from ~$L_i \in [2, 8]$ and each input item has $N_i=3$ length, and $A$ is set to $32$.
In this experiment, we decrease $A$ from $64$ to $16$ for the copy task, and from $32$ to $16$ for the association recall task, so that each memory block has less number of memory slots than the length of the input sequence.
Fig.~\ref{fig.reduced.algorithmic} shows DAM performance on the copy and the association recall task according to the number of sub memory blocks. In the results, similar to the previous experiments, the memory network performance on both tasks is enhanced as we increase the number of sub memory blocks.

\begin{figure*}[t]
\centering

\begin{subfigure}[t]{.4\textwidth}
    \centering
    \includegraphics[width=\linewidth]{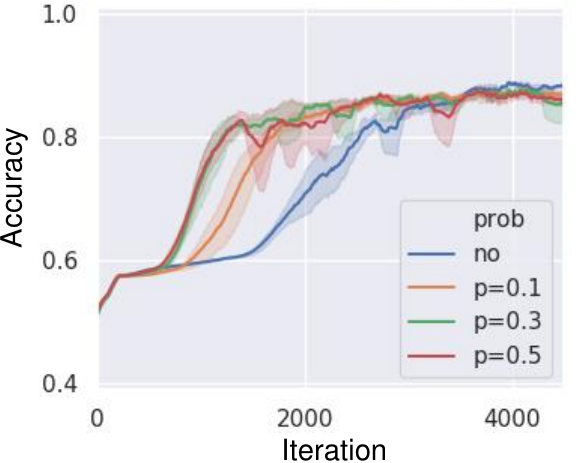}
    \caption{}
    \label{fig.mr.reduced.copy.dnc}
  \end{subfigure}
  \hspace{12pt}
  \begin{subfigure}[t]{.4\textwidth}
    \centering
    \includegraphics[width=\linewidth]{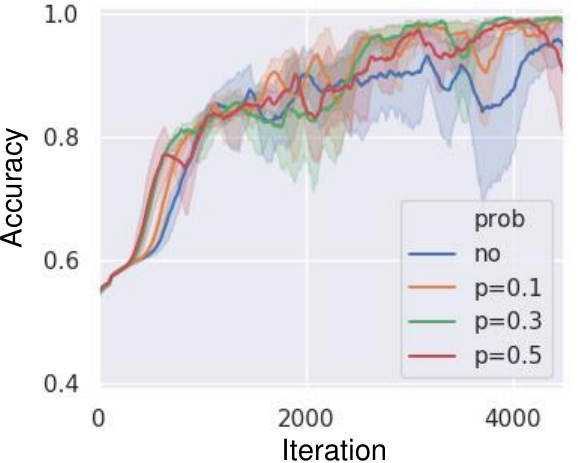}
    \caption{}
    \label{fig.mr.reduced.copy.dam2}
  \end{subfigure}
  
  \medskip
  
\begin{subfigure}[t]{.4\textwidth}
    \centering
    \includegraphics[width=\linewidth]{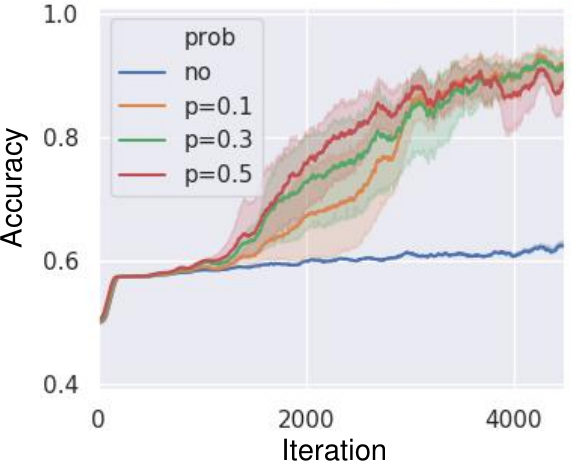}
    \caption{}
    \label{fig.mr.reduced.ar.dnc}
  \end{subfigure}
  \hspace{12pt}
  \begin{subfigure}[t]{.4\textwidth}
    \centering
    \includegraphics[width=\linewidth]{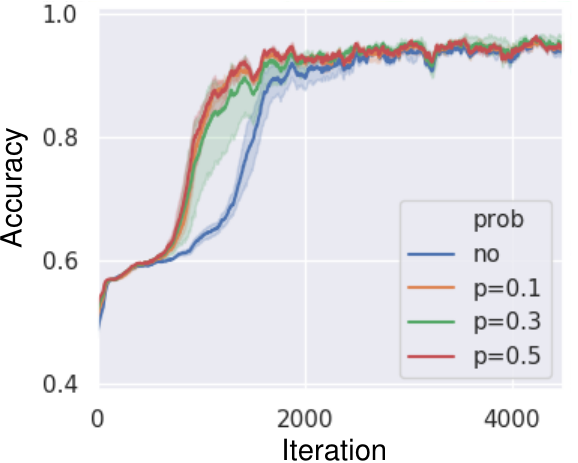}
    \caption{}
    \label{fig.mr.reduced.ar.dam2}
  \end{subfigure}
  
\caption{Mean training curves for different reproducing probability values at (a) DNC and (b) DAM-2 on the copy task with $A=16$. Mean training curves for different reproducing probability values at (c) DNC and (d) DAM-2 on the associative recall task with $A=16$. The shadowed area shows a standard deviation of 5 trials.}
\label{fig.mr.reduced.algorithmic}
\end{figure*}

Fig.~\ref{fig.mr.reduced.algorithmic} shows the experimental result for the effect of MRL on both copy task and associative recall task.
In Figs.~\ref{fig.mr.reduced.algorithmic}(a) and (b), even with the decreased number of memory slots, $A=16$, MRL still enhances the DNC and DAM performance on the copy task with increasing reproducing probability. For the associative recall task, the result shows a similar pattern as the copy task.

\section{DAM model Equations directly related to DNC}

\begin{align}
(DNC)~& \bm\xi_t = W_{\xi}\bm{h}_{t} = [W_{\xi,1}]\bm{h}_{t} \in \mathbb{R}^{L*R+3L+5R+3} \\
\begin{split}
(DAM)~& \bm\xi_t= W_{\xi} \bm{h}_{t} = [\bm\xi_{t,1},\cdots,\bm\xi_{t,K},\hat{g}_{t}^{at}] \\ &~~~~=[W_{\xi,1},\cdots,W_{\xi,K},W_{\xi,at}]\bm{h}_{t} \in \mathbb{R}^{K*(L*R+3L+3R+3)}
\end{split}
\end{align}

DNC generates the memory operators,~$\bm\xi_t$, called as interface vector, for its single memory operation, and DAM extends this vector for multiple independent memory blocks.
DAM generates $K$ number of DNC like memory operators,~$\bm\xi_{t,k}$, (except for temporal linkage operator) and newly introduce attentive gate,~$\hat{g}_{t}^{at}$ to read from those multiple memory blocks.

\begin{align}
(DNC)~& \bm{M}_t=\bm{M}_{t-1}\circ(\bm{E}-\bm{w}_t^{w}\bm{e}_t^\top)+\bm{w}_t^{w}\bm{v}_t^\top \\
(DAM)~& \bm{M}_{t,k}=\bm{M}_{t-1,k}\circ(\bm{E}-\bm{w}_{t,k}^{w}\bm{e}_{t,k}^\top)+\bm{w}_{t,k}^{w}\bm{v}_{t,k}^\top
\end{align}

The writing process of DAM is the same as DNC as shown in the above equations, except the same write operation is executed in multiple memory blocks independently at the same time.

\begin{align}
(DNC)~& \bm{r}_t = \bm{M}_t{\bm{w}_t^{r}}^\top \\
(DAM)~& \bm{r}_{t} = \sum_{k=1}^{K} g_{t,k}^{at} \bm{M}_{t,k}^\top{\bm{w}_{t,k}^{r}}
\end{align}
where $g_{t,k}^{at} = Softmax(\hat{g}_{t,k}^{at})$ for $k=1,\cdots,K$.

In the reading process of DAM, the basic reading procedure for each memory block is the same as DNC, but, DAM integrates every read-out value from $K$ memory blocks into a single distributed representation with an attentive gate. The attentive gate, $\hat{g}_{t,k}^{at}$, is a newly introduced part of DAM for the attentive interpolation.

\section{Model Implementation Details}  \label{appendix.implementation.detail}

In this section, we provide more detail of the DAM architecture including well-known neural network generalization techniques that are used in \cite{franke2018robust}.

At each time step $t$, the controller, LSTM~\cite{hochreiter1997long},  receives a input,~$\bm{x}_t \in \mathbb{R}^{d_i}$, previous hidden state,~$\bm{h}_{t-1} \in \mathbb{R}^{d_h}$, and previous memory read-out values,~$\bm{r}_{t-1} = \{\bm{r}^i_{t-1} \in \mathbb{R}^L;1 \leq i \leq R\}$, where $L$ is a memory length size and $R$ is the number of read heads.
Based on these values, the controller updates its internal state,~$\bm{h}_t = Controller([\bm{x}_t;\bm{r}_{t-1};\bm{h}_{t-1}])$.
Then, layer normalization~\cite{lei2016layer} is applied to the updated internal state.
From the normalized internal state~$\bm{h}_{t,LN}$, the controller generates a memory operators,~$\bm\xi_t \in \mathbb{R}^{K*(L*R+3L+3R+3)}$, which is called interface vector, as follows:
\begin{equation}
\label{eq.appendix.controller}
\bm\xi_t=[\bm\xi_{t,1},...,\bm\xi_{t,K},\hat{g}_{t}^{at}] =[W_{\xi,1},\cdots,W_{\xi,K},W_{\xi,at}]\bm{h}_{t,LN}
\end{equation}
where $\bm\xi_{t,k} \in \mathbb{R}^{L*R+3L+2R+3}$ is a memory operator for each memory block, $K$ is the number of memory blocks, $k \in \{1,\cdots,K\}$, and $\hat{g}_{t}^{at} \in \mathbb{R}^{K*R}$ is an attentive gate.

The interface vector, $\bm\xi_{t,k}$, is split into sub-components, and each sub-component and $\hat{g}_{t}^{at}$ are used for each memory's read/write operations as follows:

\begin{equation}
\begin{aligned}
\label{eq.appendix.interface}
\bm\xi_{t,k} = [\bm{k}_{t,k}^w;\hat{\beta_{t,k}^w};\bm{\hat{e}}_{t,k};\bm{v}_{t,k};\hat{f}_{t,k}^1,\cdots,\hat{f}_{t,k}^R;\hat{g}_{t,k}^a;\hat{g}_{t,k}^w; \\
\bm{k}_{t,k}^{r,1},\cdots,\bm{k}_{t,k}^{r,R};\hat{\beta}_{t,k}^{r,1},\cdots,\hat{\beta}_{t,k}^{r,R}]
\end{aligned}
\end{equation}

\begin{itemize}[noitemsep,topsep=0pt,parsep=5pt,partopsep=0pt, align=left]
\item the write-in \textit{key} $\bm{k}_{t,k}^w \in \mathbb{R}^L$;
\item the write strength $\beta_{t,k}^w=\zeta(\hat{\beta}_{t,k}^w) \in [1,\infty)$;
\item the erase \textit{values} $\bm{e}_{t,k}=\sigma(\bm{\hat{e}}_{t,k}) \in [0,1]^L$;
\item the write-in \textit{values} $\bm{v}_{t,k} \in \mathbb{R}^L$;
\item R free gates $\{f_{t,k}^i=\sigma(\hat{f}_{t,k}^i) \in [0,1];1 \leq i \leq R\}$;
\item the allocation gate $g_{t,k}^a=\sigma(\hat{g}_{t,k}^a) \in [0,1]$;
\item the write gate $g_{t,k}^w=\sigma(\hat{g}_{t,k}^w) \in [0,1]$;
\item the read-out \textit{keys} $\{\bm{k}_{t,k}^{r,i} \in \mathbb{R}^L; 1 \leq i \leq R\}$;
\item the read strengths $\{\beta_{t,b}^{k,i}=\zeta(\hat{\beta}_{t,k}^{r,i}) \in [1,\infty) ;1 \leq i \leq R\}$; and
\item the attentive gate $\{g_{t,k}^{at,i} = Softmax(\hat{g}_{t,k}^{at,i})~~~\mbox{for}~k=1,\cdots,K.~;1 \leq i \leq R\}$.
\end{itemize}
where $\zeta(\cdot)$ denotes a Oneplus function, $\sigma(\cdot)$ denotes a Sigmoid function and $Softmax(\cdot)$ denotes a Softmax function.

Based on those memory operators, the model performs a writing process for each memory block simultaneously.
The controller finds a writing address,~$\bm{w}^w_{t,k} \in [0,1]^A$, where $A$ is a memory address size, in two ways: (i) a content-based addressing and (ii) a memory usage statistic.

The content-based addressing computes data address as following:
\begin{equation}
\label{eq.appendix.contents}
\mathcal{C}(\bm{M},\bm{k},\beta)[i] = \frac{exp\{\mathcal{D}(\bm{k},\bm{M}[i,\cdot])\beta\}}{\sum_{j}exp\{\mathcal{D}(\bm{k},\bm{M}[j,\cdot])\beta\}}
\end{equation}
where $\mathcal{D}(\cdot,\cdot)$ is the cosine similarity, $\bm{M} \in \mathbb{R}^{A \times L}$ is the memory matrix, and the $\beta$ controls a strength of the address's sharpness.

Therefore, It finds write-content addresses,~$\bm{c}_{t,b}^w \in [0,1]^A$, based on the cosine similarity between write-in \textit{keys} and memory values, as follows:
\begin{equation}
\label{eq.appendix.write.contents}
\bm{c}_{t,k}^w=\mathcal{C}(\bm{M}_{t,k},\bm{k}_{t,k}^w,\beta_{t,k}^w)
\end{equation}

Next, it finds an allocation address,~$\bm{a}_{t,k} \in [0,1]^A$, considering current memory usage, such as unused or already read memory space.
It determines the retention of the most recently read address's values for each memory block through a memory retention vector,~$\bm\psi_{t,k} \in [0,1]^A$ and it calculates current memory usage vector,~$\bm{u}_{t,k} \in [0,1]^A$ as follows:
\begin{equation}
\label{eq.appendix.retention}
\bm\psi_{t,k} = \prod_{i=1}^{R} (\bm{1}-f_{t,k}^i \bm{w}^{r,i}_{t-1,k})
\end{equation}
\begin{equation}
\label{eq.appendix.usage}
\bm{u}_{t,k} = (\bm{u}_{t-1,k} + \bm{w}^{w}_{t-1,k} - \bm{u}_{t-1,k} \circ \bm{w}^{w}_{t-1,k}) \circ \bm\psi_{t,k}
\end{equation}
where $\circ$ denotes a element-wise multiplication.

Afterwards, $\bm{a}_{t,k}$ is determined based on current memory usage information as follows:
\begin{equation}
\label{eq.appendix.allocation}
\bm{a}_{t,k}[\phi_{t,k}[j]] = (1 - \bm{u}_{t,k}[\phi_{t,k}[j]]) \prod_{i=1}^{j-1} \bm{u}_{t,k}[\phi_{t,k}[i]]
\end{equation}
where $\phi_{t,k}$ is a free list which informs indices sorted with respect to memory usage values, i.e. $\phi_{t,k}[1]$ is the index of the least used address.

The writing address for each memory block,~$\bm{w}^{w}_{t,k}$, is determined by interpolation between $\bm{c}_{t,k}^w$ and $\bm{a}_{t,k}$, and each memory block is updated, as follows: 
\begin{equation}
\label{eq.appendix.writing.address}
\bm{w}_{t,k}^w = g_{t,k}^w[g_{t,k}^a\bm{a}_{t,k}+(1-g_{t,k}^a)\bm{c}_{t,k}^w]
\end{equation}
\begin{equation}
\label{eq.appendix.write.memory}
\bm{M}_{t,k}=\bm{M}_{t-1,k}\circ(\bm{E}-\bm{w}_{t,k}^{w}\bm{e}_{t,k}^\top)+\bm{w}_{t,k}^{w}\bm{v}_{t,k}^\top
\end{equation}
where $g_{t,k}^a$ controls a proportion between $\bm{a}_{t,k}$ and $\bm{c}_{t,k}^w$, $g_{t,k}^w$ controls an intensity of writing, $\circ$ denotes element-wise multiplication and $E$ is $\bm{1}^{A \times L}$.

After all memory blocks are updated, the model executes a reading process for each memory block.
It finds out a read address for each memory block,~$\bm{w}_{t,k}^{r,i} \in [0,1]^A$, as follows:
\begin{equation}
\label{eq.appendix.reading.address}
\bm{w}_{t,k}^{r,i}=\mathcal{C}(\bm{M}_{t,k},\bm{k}_{t,k}^{r,i},\beta_{t,k}^{r,i})
\end{equation}
where $i \in \{1,...,R\}$.

Then, it reads preliminary read-out value,~$\bm{r}_{t,k}^{i} \in \mathbb{R}^L$, from each memory block and interpolates the preliminary read-out values by $g_t^{at}$ to produce read-out value,~$\bm{r}_{t}^{i} \in \mathbb{R}^L$, as follows:
\begin{equation}
\label{eq.appendix.preliminary.read.value}
\bm{r}_{t,k}^{i} = \bm{M}_{t,k}^\top{\bm{w}_{t,k}^{r,i}}
\end{equation}
\begin{equation}
\label{eq.appendix.read.value}
\bm{r}_t^{i} = \sum_{k=1}^{K} g_{t,k}^{at,i} \bm{r}_{t,k}^{i}
\end{equation}
where $g_{t,k}^{at,i}$ controls which memory block should be used for final read-out value.

The read-out values,~$\bm{r}^{t} = \{\bm{r}^i_t ; 1 \leq i \leq R\}$, are provided to the controller.
As in \cite{franke2018robust}, the model applies drop-out~\cite{srivastava2014dropout} to $\bm{h}_{t,LN}$ with a drop-out probability,~$p_{dp}$, which is computed as $\bm{h}_{t,dp}=DropOut(\bm{h}_{t,LN} \vert p_{dp})$.
Finally, the controller produces an output,~$\bm{y}_t \in \mathbb{R}^{d_o}$, along with $\bm{h}_{t,dp}$ and $\bm{r}^t$, as follows:
\begin{equation}
\label{eq.appendix.output}
\bm{y}_t=W_y[\bm{h}_{t,dp};\bm{r}^t]
\end{equation}

\end{document}